%% file: main.tex
\documentclass[10pt,twocolumn,letterpaper]{article}

\usepackage[pagenumbers]{cvpr} 
\definecolor{cvprblue}{rgb}{0.21,0.49,0.74}
\usepackage[pagebackref,breaklinks,colorlinks,citecolor=cvprblue]{hyperref}

\usepackage{amsmath}
\usepackage{amssymb}
\usepackage{bbding}
\usepackage{mathtools}
\usepackage{amsthm}
\usepackage[ruled,vlined,linesnumbered]{algorithm2e}
\usepackage{multirow}
\usepackage{xcolor,colortbl}

\usepackage[capitalize,noabbrev]{cleveref}
\crefname{section}{Sec.}{Secs.}
\Crefname{section}{Section}{Sections}
\Crefname{table}{Table}{Tables}
\crefname{table}{Tab.}{Tabs.}
\crefname{figure}{Fig.}{Figs.}
\crefname{wrapfigure}{Fig.}{Figs.}
\crefname{equation}{Eq.}{Eqs.}
\crefname{algorithm}{Alg.}{Algs.}
\crefname{appendix}{App.}{Apps.}
\SetCommentSty{myCommentStyle}

\usepackage{enumitem}
\setenumerate[1]{itemsep=0pt,partopsep=0pt,parsep=\parskip,topsep=5pt}
\setitemize[1]{itemsep=0pt,partopsep=0pt,parsep=\parskip,topsep=5pt}
\setdescription{itemsep=0pt,partopsep=0pt,parsep=\parskip,topsep=5pt}

\def\paperID{***} 
\def\confName{CVPR}
\def\confYear{2025}

\title{Unlocking Tuning-Free Few-Shot Adaptability\\in Visual Foundation Models by Recycling Pre-Tuned LoRAs}

\author{Zixuan Hu\\
Nanyang Technological University\\
{\tt\small ZIXUAN014@e.ntu.edu.sg}
\and
Yongxian Wei\\
Tsinghua University\\
{\tt\small weiyx23@mails.tsinghua.edu.cn}
\and
Li Shen\\
Sun Yat-sen University\\
{\tt\small mathshenli@gmail.com}
\and
Chun Yuan\\
Tsinghua University\\
{\tt\small yuanc@sz.tsinghua.edu.cn}
\and
Dacheng Tao\\
Nanyang Technological University\\
{\tt\small dacheng.tao@gmail.com}
}

\begin{document}
\maketitle
\input{sec/0_abstract}    
\input{sec/1_intro}

\input{sec/2_related}

\input{sec/3_methodology}
\input{sec/4_experiment}

\section{Conclusion}
In this paper, we propose LoRA Recycle, a novel meta-learning framework to achieve tuning-free few-shot adaptation in VFMs, by reusing diverse pre-tuned LoRAs without access to original training data. 
 We further propose the double-efficient mechanism tailored to our framework, significantly accelerating the meta-training by selectively using the most informative tokens, while maintaining or even  improving performance by reducing noise.
 Experimental results across various few-shot classification benchmarks, across in-domain and challenging cross-domain scenarios, confirm the effectiveness of LoRA Recycle.

\nocite{lin2024training}
{
    \small
    \bibliographystyle{ieeenat_fullname}
    \bibliography{main}
}

\input{sec/sup}

\end{document}

%% file: sec/0_abstract.tex
\begin{abstract}
Large Language Models (LLMs) such as ChatGPT demonstrate strong few-shot adaptability without requiring fine-tuning, positioning them ideal for data-limited and real-time applications. However, this adaptability has not yet been replicated in current Visual Foundation Models (VFMs), which require explicit fine-tuning with sufficient tuning data.
Besides, the pretraining-finetuning paradigm has led to the surge of numerous task-specific modular components, such as Low-Rank Adaptation (LoRA).
For the first time, we explore the potential of reusing diverse pre-tuned LoRAs without accessing their original training data, to achieve tuning-free few-shot adaptation in VFMs.
Our framework, LoRA Recycle,  distills a meta-LoRA from diverse pre-tuned LoRAs with a meta-learning objective, using surrogate data generated inversely from pre-tuned LoRAs themselves.
The VFM, once equipped with the meta-LoRA, is empowered to solve new few-shot tasks in a single forward pass, akin to the in-context learning of LLMs.
Additionally, we incorporate a double-efficient mechanism tailored to our framework, significantly accelerating the meta-training process while maintaining or even improving performance. Extensive experiments across various few-shot classification benchmarks across both in- and cross-domain scenarios demonstrate the superiority of our framework.

\end{abstract}

%% file: sec/1_intro.tex
\section{Introduction}
\label{sec:intro}
Large Language Models (LLMs) like ChatGPT demonstrate a profound capacity to solve few-shot tasks without the necessity for fine-tuning, making them ideal for data-limited and real-time applications.
However, this adaptability can not be replicated by current Visual Foundation Models (VFMs), which typically require explicit fine-tuning with sufficient tuning data.

Low-Rank Adaptation (LoRA) \citep{hu2021LoRA} has emerged as a prominent fine-tuning approach in current research, valued for its capacity to achieve strong fine-tuning performance with sufficient tuning data, while only updating a small subset of additional parameters—specifically, trainable rank decomposition matrices—rather than the entire model.
While promising, (i) explicit fine-tuning is often prohibitive for  applications requiring real-time responses, and (ii) fine-tuning with limited data is extremely unstable.
As shown in \cref{tab:motivation}, fine-tuning with limited data makes performance highly sensitive to choices like the optimizer, learning rate, and step size. Besides, it introduces unacceptable latency for applications requiring real-time responses.

\begin{figure}[!t]
\vspace{-0.1cm}
    \centering
    \includegraphics[width=1.0\linewidth]{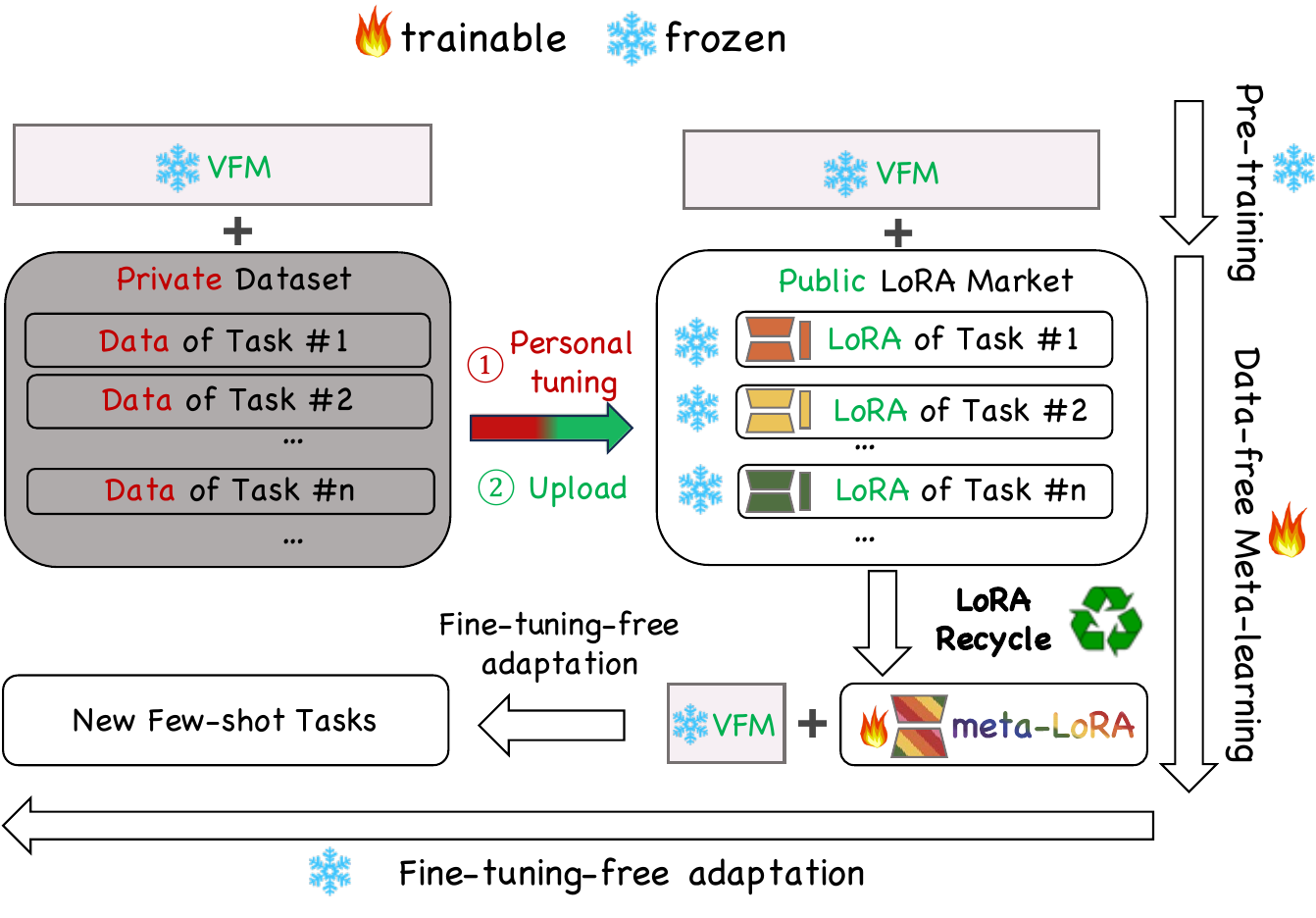}
\vspace{-0.7cm}
  \caption{ Thanks to the modularity of LoRA, users can upload locally tuned LoRAs to public repositories without exposing original training data.
  LoRA Recycle distills a meta-LoRA from these LoRAs without needing their original training data. The VFM, once equipped with the meta-LoRA, is empowered to solve new few-shot tasks in a single forward pass without fine-tuning.}
  \label{fig:motivation}
\vspace{-0.31cm}
\end{figure}

For the first time, we explore the potential of reusing diverse pre-tuned LoRAs without accessing their original training data, 
to achieve tuning-free few-shot adaptation in VFMs (see \cref{fig:motivation}).
Our inspiration comes from the concept of \textit{LoRA Market} \citep{huang2023LoRAhub}, where diverse pre-tuned LoRAs are publicly accessible without exposing original training data due to privacy concerns.  For task-specific reuse, users can download and then insert a task-specific LoRA of interest into the open-source VFM, to obtain a personalized VFM.
Moving beyond task-specific reuse, we seek to leverage the vast availability and diversity of these LoRAs from a novel perspective, leading to our central research question: 
\textit{
Is it feasible to reuse diverse pre-tuned LoRAs without accessing their original training data, to achieve tuning-fee few-shot adaptation in VFMs?}
This offers new insights into leveraging the vast accessibility and diversity of LoRAs beyond task-specific reuse, and avoids the need to access original training data, which is often restricted by privacy concerns.

To answer this question, we propose a meta-learning framework named \textbf{LoRA Recycle} (see \cref{fig:pipeline}).
Without access to the original training data, we propose LoRA Inversion to generate surrogate data inversely from the pre-tuned LoRAs themselves. A meta-LoRA is then distilled from the pre-tuned LoRAs using these surrogate data, with a meta-learning objective of learning how to adapt to diverse tasks without fine-tuning. Thanks to the meta-learning objective, the VFM, once equipped with the meta-LoRA, is empowered to adapt to new few-shot tasks in a single forward pass without further fine-tuning, akin to in-context learning of LLMs. 
The core idea of LoRA Recycle is to reshape the VFMs' prior over a distribution of expected tasks (\ie, pre-tuned LoRAs) via meta-learning, and such prior encoded in the meta-LoRA can facilitate learning of new tasks sampled from similar distributions. 
To further improve efficiency, we introduce a double-efficient mechanism tailored to our framework. During the inversion stage, it prunes unimportant tokens based on self-attention weights in hidden layers, thereby accelerating data generation. The pruning results from the inversion stage further help to indicate the informative and sparse tokens in the generated data, which are then exclusively used in the subsequent meta-training stage. This selective use of sparse tokens significantly accelerates the meta-training process, while maintaining or even  improving performance by reducing noise from generated data.
We summarize our contributions as follows:
\begin{itemize}[itemsep=5pt, parsep=-5pt,leftmargin=8pt]
    \item \textbf{Novel perspective:} We are the first to enable tuning-free few-shot adaptation in VFMs from the perspective of LoRA reusing, offering new insights into leveraging the accessibility and diversity of pre-tuned LoRAs beyond traditional task-specific reuse.
    \item \textbf{United framework:} (i) We propose a meta-learning framework named LoRA Recycle, effectively achieving tuning-free few-shot adaptation in VFMs by reusing diverse pre-tuned LoRAs without needing their original training data. (ii) We further propose a double-efficient mechanism tailored to our framework, which significantly accelerates the meta-training process by selectively using the most informative tokens, while maintaining or even improving performance by reducing noise.
    \item \textbf{Experiments:} Extensive experiments across various few-shot classification benchmarks, within both in-domain and cross-domain scenarios, demonstrate the effectiveness of LoRA Recycle. Notably, LoRA Recycle achieves an average 6.27\% improvement over baselines in the 5-way 1-shot in-domain setting.
\end{itemize}

 \begin{table}[!t]
     \centering
     \caption{ Fine-tuning ViT-B/16 on 600 5-way 1-shot classification tasks from the meta-testing set of CIFAR-FS. We report the accuracy, throughput (tasks per second) and GPU memory usage during fine-tuning. Values highlighted in \textcolor{green}{green} represent the best, whereas those in \textcolor{red}{red} denote the worst.} 
  \scalebox{0.56}{\begin{tabular}{cccccccc}
    \toprule
     \multirow{2}{*}{\textbf{Method}} &\multirow{2}{*}{\textbf{Optimizer}}&\multirow{2}{*}{\textbf{Step}}& \multicolumn{3}{c}{\textbf{Learning Rate}}&\multirow{2}{*}{\textbf{\shortstack{Throughput\\(tasks/s) \textcolor{blue}{$\uparrow$}}}}&\multirow{2}{*}{\textbf{\shortstack{GPU Mem\\(GB) \textcolor{blue}{$\downarrow$}}}}\\
    \cmidrule(r){4-6}
     &&&{{0.1}}&{{0.01}}&{{0.001}}&&\\
     \midrule
     \multirow{4}{*}{\shortstack{Full\\Fine-Tuning}}&\multirow{2}{*}{SGD} &50 & 22.81  & \textcolor{green}{30.13} & 28.99 &\multirow{4}{*}{0.10}  &\multirow{4}{*}{12.88}\\
     & &5   &  20.56 & 23.69& 23.85&&\\
     \cmidrule(r){3-6}
     &\multirow{2}{*}{Adam} &50   & 20.04  & 20.43& 26.64&&\\
     & &5    & 20.00  &\textcolor{red}{19.96} & 21.09&&\\
    \midrule
     \multirow{4}{*}{LoRA}&\multirow{2}{*}{SGD} &50 & {79.29}  & 77.07 & {36.61}&\multirow{4}{*}{0.13} & \multirow{4}{*}{9.54}\\
     & &5   & {73.48}  &37.27 & {20.60}&&\\
     \cmidrule(r){3-6}
     &\multirow{2}{*}{Adam} &50   &  {22.10} &26.55 & \textcolor{green}{82.19}&&\\
     & &5    &  \textcolor{red}{20.40} & {73.00}& 55.11&&\\
     \midrule
     {\shortstack{\textbf{LoRA} \textbf{Recycle (ours)}}} & {---} & {---} & \multicolumn{3}{c}{{\textbf{89.70} \textcolor{blue}{(+7.51\%)}}} & {\textbf{8.25} \textcolor{blue}{($\times$63)}} &{\textbf{1.28} \textcolor{blue}{(-87\%)}}\\
     \bottomrule
  \end{tabular}}
     \vspace{-0.3cm}
  \label{tab:motivation}
 \end{table}

%% file: sec/2_related.tex
\section{Related Work}
\label{sec:relatedWork}

\subsection{LoRA \& LoRA Reuse}
Fine-tuning the entire foundation model results in high costs in computation and storage. 
To mitigate these challenges, several parameter-efficient fine-tuning (PEFT) methods \citep{hu2021LoRA,he2022towards,wu2023pi,liu2022few,he2021towards,jiang2023rethinking} have emerged, focusing on the update of a limited subset of model parameters.
LoRA \citep{hu2021LoRA}  parallelly attaches extra low-rank decomposition matrices  to original weights, while adapter tuning \citep{houlsby2019parameter,gao2023clip,bansal2022meta} sequentially appends extra layers  behind the original feed-forward layers. 
More recently, several works \citep{huang2023LoRAhub,wu2023pi,gou2023mixture,chen2024llava,wu2023mole} have investigated the potential of composing multiple pre-tuned LoRAs. However, (i) they are limited to parameter arithmetic like weight averaging, lacking
 precise alignment for LoRAs targeting different label spaces in the context of classification. (ii) They are not specifically designed to achieve tuning-free few-shot adaptation in VFMs. (iii) They are not applicable to reuse LoRAs with different architectures like different ranks.


\subsection{Meta-Learning \& Data-Free Meta-Learning}
Meta-learning, also known as \textit{learning to learn}, aims to 
learn prior knowledge over a distribution of tasks, enabling efficient adaptation to unseen few-shot tasks from similar distributions.
Data-based meta-learning \citep{finn2017model,yoon2018bayesian,khodak2019adaptive,fu2023styleadv} typically assumes the availability of task-specific data for each meta-training task. Recently, Data-Free Meta-Learning (DFML) \citep{wang2021meta,hu2023architecture, hu2023learning,wei2024free,wei2024task} emerges as a promising solution to directly meta-learn from pre-trained models available off the shelf.
However, existing methods struggle to scale up to large Vision Transformers, whereas our framework only meta-trains a lightweight meta-LoRA. Furthermore, we introduce an innovative double-efficient mechanism that significantly accelerates the meta-training process by selectively using the most informative tokens.

\subsection{Tuning-Free Adaptation of Foundation Models}
Compared to explicit fine-tuning, training-free adaptation requires no parameter updates, making it highly suitable for real-time applications with low computational budgets. LLMs achieve tuning-free adaptation through their inherent in-context learning capabilities \citep{dong2022survey}. Existing studies suggest that in-context learning is equivalent to implicitly performing gradient descent \citep{dai2022can, von2023transformers}, viewing LLMs as meta-learning models \citep{NEURIPS2020_1457c0d6}. However, this in-context learning ability has not yet been replicated by current VFMs. To address this, \cite{fifty2023context} explicitly trains a sequence model with VFMs to simulate LLM-style in-context learning. \cite{zhang2024personalize,liu2024matcher} adapt the Segment Anything Model in a tuning-free manner using a one-shot example.
Our LoRA Recycle, on the other hand, achieves tuning-free adaptation to few-shot tasks from a novel perspective, by reusing diverse pre-tuned LoRAs without needing their original training data.

%% file: sec/3_methodology.tex
\section{Preliminary \& Problem Setup}
\label{sec:problem}
\textbf{Low-Rank Adaptation (LoRA)} \citep{hu2021LoRA} enables VFM to solve a specific task by only updating lightweight extra modules. For a weight matrix $W^{(l)} \in \mathbb{R}^{d \times k}$ at the $l^{th}$ layer within the VFM $f$, a LoRA module is represented as a low-rank matrix decomposition $\delta{W}^{(l)}=\delta{W}_{\rm A}^{(l)}\cdot \delta{W}_{\rm B}^{(l)}$, where $\delta{W}_{\rm A}^{(l)} \in \mathbb{R}^{d \times r}$, $\delta{W}_{\rm B}^{(l)} \in \mathbb{R}^{r \times k}$ and the rank $r \ll \min(d,k)$. The input $\mathbf{X}_{\rm in}$ will be processed in parallel as $\mathbf{X}_{\rm out}^{(l)}=W^{(l)}\mathbf{X}_{\rm in}^{(l)} + \delta{W}_{\rm A}^{(l)} \delta{W}_{\rm B}^{(l)}\mathbf{X}_{\rm in}^{(l)}$. When fine-tuning, it freezes the original weight matrix $W$ while only keeping $\delta{W}_{\rm A}$ and $\delta{W}_{\rm B}$ trainable. 
When facing classification tasks, a classification head $h$ is always tuned together with the LoRA modules to output the prediction distribution. We use $f_{\delta{W}}$ to denote the VFM equipped with the LoRA $\delta{W}$.


\noindent
\textbf{Problem setup: LoRA Recycle.}\ We are given a transformer-based VFM $f$ pre-trained on large-scale datasets, and multiple LoRAs with classification heads pre-tuned on diverse classification tasks. Following standard meta-learning
setup \citep{finn2017model}, we assume these tasks follow an underlying task distribution $p_{\mathcal{T}}$. $(\delta{W}_{\mathcal{T}}, h_{\mathcal{T}}) \sim p_{\mathcal{T}}$ denotes the LoRA and classification head pre-tuned on task $\mathcal{T}$. Note that we have no access to the original training data behind the given LoRAs.
Our goal is to meta-train a meta-LoRA $\delta{W}^{*}$ over $p_{\mathcal{T}}$, so that the VFM $f$, once equipped with $\delta{W}^{*}$ (\textit{i.e.}, $f_{\delta{W}^{*}}$), can adapt to new few-shot tasks sampled from similar distributions without further fine-tuning.

\noindent
\textbf{Testing setup.}\ We conduct evaluation on 600 $N$-way $K$-shot classification tasks. Note that the classes in these testing tasks have not been seen by any given LoRA. Each $N$-way $K$-shot task $\mathcal{T}$ consists of one support set $\mathcal{D}_{\rm s}^{\mathcal{T}}$ and one query set $\mathcal{D}_{\rm q}^{\mathcal{T}}$.  The support set $\mathcal{D}_{s}^{\mathcal{T}}$ has $N$ classes and $K$ examples per class. We focus on a few-shot setting where $K$ is small (e.g., 1 or 5), thus fine-tuning $f$ with extremely few examples is infeasible. In contrast, we use $\mathcal{D}_{\rm s}^{\mathcal{T}}$  to adapt $f$ in a tuning-free manner. The query set $\mathcal{D}_{\rm q}^{\mathcal{T}}$ is what we actually make predictions on.  The overall accuracy is measured by averaging the accuracy across all testing tasks.

\begin{figure*}[t]
  \centering
    \includegraphics[width=0.95\linewidth]{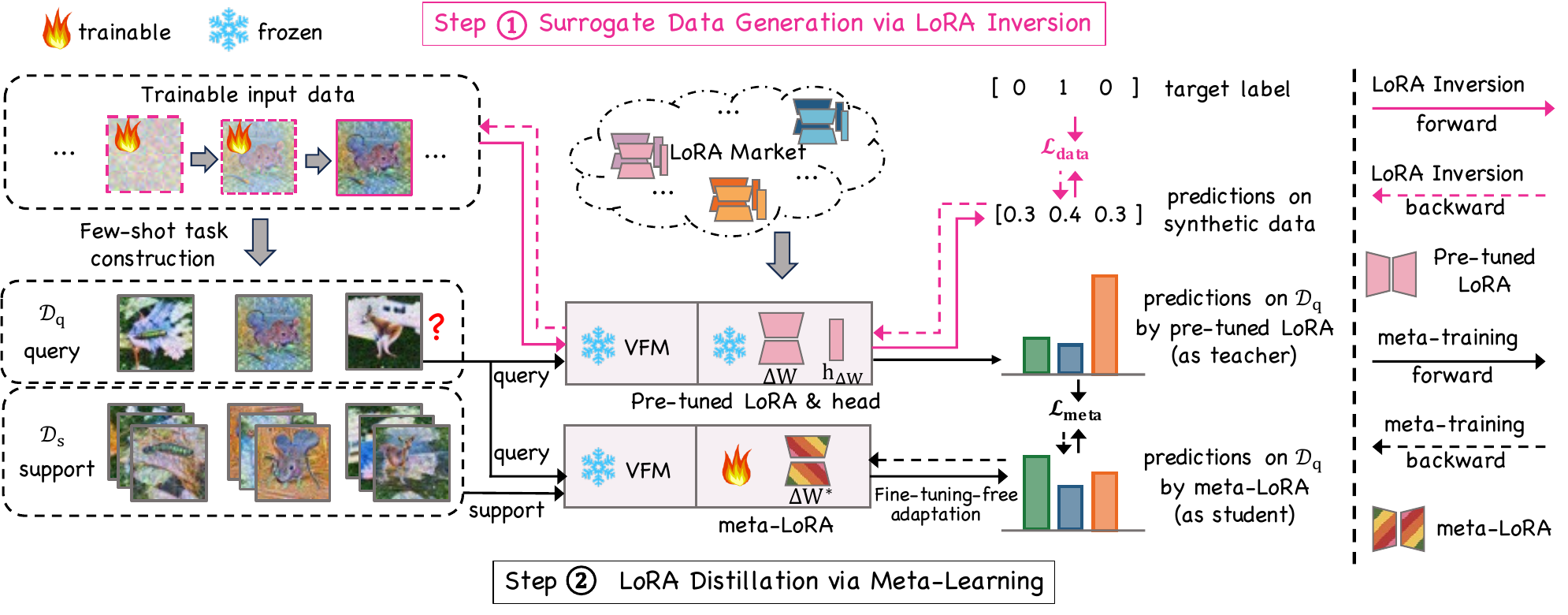}
   \vspace{-0.1cm}
   \caption{Pipeline of LoRA Recycle. \textbf{(i) (Pink Path)} We generate task-specific surrogate data from the pre-tuned LoRA via LoRA Inversion. The input data (attached with the fire in the left corner) is initialized as Gaussian noise and iteratively optimized by minimizing $\mathcal{L}_{\rm data}$ (\cref{eq:inversion}). The surrogate data is then used to construct a meta-training task with one support set and one query set. \textbf{(ii) (Black Path)} We meta-train the meta-LoRA (attached with the fire in the middle) on a  wide range of pre-tuned LoRAs by minimizing the meta-learning objective $\mathcal{L}_{\rm meta}$ (\cref{eq:LoRAhubDistillation}), explicitly teaching it how to adapt without fine-tuning.
   }
   \label{fig:pipeline}
   \vspace{-0.3cm}
\end{figure*}

\section{Methodology}
\label{sec:method}
In this section, we present our proposed framework LoRA Recycle (see \cref{fig:pipeline} and \cref{alg:training}).
Without access to original training data, we propose LoRA Inversion to generate surrogate data from pre-tuned LoRAs (see \cref{sec:loraInversion}). A meta-LoRA is then distilled from the pre-tuned LoRAs using these surrogate data, with a meta-learning objective of learning how to adapt without fine-tuning (see \cref{sec:lorahubDistillation}).
To further improve efficiency, we propose a double-efficient mechanism significantly accelerating the meta-training process by selectively using the most informative tokens, while also improving performance (\cref{sec:double}) by reducing noise from the generated data.

\subsection{Surrogate Data Generation via LoRA Inversion}
\label{sec:loraInversion}
\textbf{LoRA Inversion.} Given a pre-tuned LoRA $\delta{W}$ with its classification head $h$, we generate its original training data by iteratively optimizing (a batch of) data $\mathbf{X}$, which is initialized as Gaussian noise. This is done by minimizing the following loss function:
\begin{equation}
    \min_{\mathbf{X}} \mathcal{L}_{\rm data}= \text{CE}\left(h\circ f_{\delta{W}}(\mathbf{X}),\mathbf{Y}\right) + \alpha_{\mathcal{R}}\mathcal{R}_{\rm BN}(\mathbf{X}),\label{eq:inversion}
\end{equation}
where $\mathbf{Y}$ is the target label (\textit{e.g.}, $[1, 0, 0]$). ${\rm CE(\cdot)}$ is a cross-entropy classification loss. $\mathcal{R}_{\rm BN}$ is an image regularization term with a coefficient $\alpha_{\mathcal{R}}$. Minimizing the first classification loss is to achieve label-conditional generation, ensuring $\mathbf{X}$ can be predicted by $f_{\delta{W}}$ as the target label $\mathbf{Y}$. To further improve the realism of the generated data, we impose a naturalness prior $\mathcal{R}_{\rm BN}$ \citep{yin2020dreaming}:
\begin{equation}
    \mathcal{R}_{\rm BN}(\mathbf{X})=\sum_l\left\|\mu^{(l)}(\mathbf{X})-\mu_{\mathrm{BN}}^{(l)}\right\|_2+\left\|\sigma^{(l)}(\mathbf{X})-\sigma_{\mathrm{BN}}^{(l)}\right\|_2,
    \label{eq:bn}
\end{equation}
where $\mu^{(l)}(\mathbf{X})$ and $\sigma^{(l)}(\mathbf{X})$ denote the mean and variance of the inputs' feature maps calculated at the $l^{\rm th}$ layer of the pre-trained model. $\mu_{\rm BN}^{(l)}$ and $\sigma_{\rm BN}^{(l)}$ denote the statistics initially stored in the $l^{th}$ batch normalization (BN) layer of the pre-trained model, which is calculated with the original training data. Given that Vision Transformers do not have the BN layer, \cite{hatamizadeh2022gradvit} suggest that we can borrow the BN statistics stored in an open-source pre-trained ResNet50. Since  $\mu_{\rm BN}^{(l)}$ and $\sigma_{\rm BN}^{(l)}$ is calculated with real data, minimizing gaps in these statistics can align the distribution between the generated and real data, thus improving realism (see \cref{fig:bn}).

\noindent
\textbf{Meta-training task construction.} After generating task-specific data of task $\mathcal{T}$, we construct a few-shot task by splitting the generated data into one support set $\mathcal{D}^{\mathcal{T}}_{\rm c}$ and one query set $\mathcal{D}^{\mathcal{T}}_{\rm q}$. This constructed task serves as a meta-training task \cite{finn2017model} for the subsequent meta-training process. In an $N$-way $K$-shot setup, the support set has $N$ classes and $K$ examples per class, while the query set  has the same $N$ classes but more examples per class, typically 15.

\subsection{LoRA Distillation via Meta-Learning}
\label{sec:lorahubDistillation}
\textbf{Meta-learning objective.} We distill a meta-LoRA $\delta{W}^*$ from diverse pre-tuned LoRAs using the surrogate data.
The meta-learning objective is formulated as follows:
\begin{subequations}
\small
\begin{align}
\small
    &\min_{\delta{W}^*} \mathcal{L}_{\rm meta}= \nonumber\\
    & \mathbb{E}_{ p_{\mathcal{T}}} \sum_{(\mathbf{X}_{\rm q},\mathbf{Y}_{\rm q})\in \mathcal{D}_{\rm q}^{\mathcal{T}}} \text{KL}\left(P(\mathbf{Y}_{\rm pred} | \mathbf{X}_{\rm q}, \mathcal{D}_{s}^{\mathcal{T}}),h_{{\mathcal{T}}}\circ f_{\delta{W}_{\mathcal{T}}}(\mathbf{X}_{\rm q})\right),  \label{eq:LoRAhubDistillation1}\\
\small
    &\text{where,} \quad P(\mathbf{}_{\rm pred}=i | \mathbf{X}_{\rm q}, \mathcal{D}_{\rm s}^{\mathcal{T}})=\frac{\exp \left(-\Vert f_{\delta{W}^*}(\mathbf{X}_{\rm q}) - \boldsymbol{c}_i \Vert_2\right)}{\sum_{i^{\prime}} \exp \left(-\Vert f_{\delta{W}^*}(\mathbf{X}_{\rm q}) - \boldsymbol{c}_{i^{\prime}} \Vert_2\right)}. \label{eq:LoRAhubDistillation2}
    \end{align}
\label{eq:LoRAhubDistillation}
\end{subequations}
Here, $p_{\mathcal{T}}$ is the underlying task distribution. $(\delta{W}_{\mathcal{T}}, h_{\mathcal{T}}, \mathcal{D}_{\rm s}^{\mathcal{T}}, \mathcal{D}_{\rm q}^{\mathcal{T}})$ refer to the pre-tuned LoRA, classification head, generated support set, and query set of task $\mathcal{T}$, which can be viewed as sampling from the task distribution $p_{\mathcal{T}}$.
The optimization in \cref{eq:LoRAhubDistillation} involves one inner loop \cref{eq:LoRAhubDistillation2} and one outer loop \cref{eq:LoRAhubDistillation1}.

\begin{itemize}[itemsep=3pt, parsep=-3pt,leftmargin=7pt]
    \item \textbf{Inner Loop:} We recast the inner loop \cref{eq:LoRAhubDistillation2} as a tuning-free adaptation: we use the support set to calculate the class center $\boldsymbol{c}_i$ of each class $i$ as the average feature embedding ($\boldsymbol{c}_i = \frac{1}{|\mathcal{D}^{\mathcal{T}}_{{\rm s}, i}|}\sum_{\mathbf{X} \in \mathcal{D}^{\mathcal{T}}_{{\rm s}, i}}f_{\delta{W}^*}(\mathbf{X})$). We then model the probability of a query example $\mathbf{X}_{\rm q} \in \mathcal{D}_{\rm q}^{\mathcal{T}}$ belonging to a class based on its Euclidean distance to the corresponding class center.
This process does not involve any parameter updating, thus avoiding calculating any second-order derivatives \citep{nichol2018first}.
    \item \textbf{Outer Loop:} In the outer loop \cref{eq:LoRAhubDistillation1}, we optimize the meta-LoRA $\delta{W}^{*}$ so that it can make more accurate predictions in the inner loop across diverse tasks. Specifically, we minimize the prediction disagreements (\textit{i.e.}, the Kullback-Leibler (KL) divergence) on the query set $\mathcal{D}_{\rm q}^{\mathcal{T}}$ between the pre-tuned LoRA $\delta{W}_{\mathcal{T}}$ (as the teacher) and the meta-LoRA $\delta{W}^{*}$ (as the student). 
The meta-LoRA $\delta{W}^{*}$ is meta-trained  across a wide range of pre-tuned LoRAs sampled from $p_{\mathcal{T}}$, explicitly learning to how to solve diverse tasks without fine-tuning.
\end{itemize}

\begin{figure*}[!t]
    \centering
\includegraphics[width=1\linewidth]{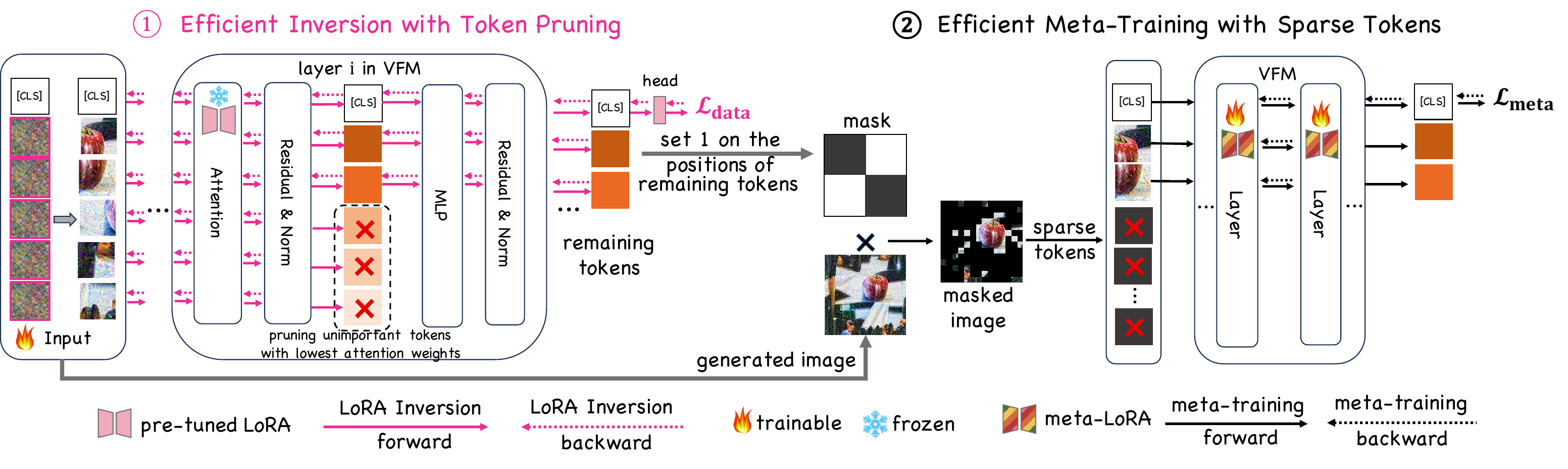}
   \vspace{-0.7cm}
   \caption{Double-Efficient Mechanism. \textbf{(Left: Inversion Stage)} During the inversion stage, token pruning is applied in the hidden layers by removing unimportant tokens based on self-attention weights, accelerating both forward and backward computations for data generation. \textbf{(Right: Meta-Training Stage)} To highlight the most informative areas in the generated image, we construct a mask by setting values of 1 at the positions of remaining tokens and 0 elsewhere. We multiply the mask with the generated image to create a masked image. We then exclusively use the unmasked tokens for the following meta-training stage. This selective use of sparse tokens significantly accelerates the meta-training process, while maintaining or even  improving performance by reducing noise from the generated data.}
   \label{fig:double}
   \vspace{-0.01cm}
\end{figure*}

\noindent
\textbf{Cross-task interpolation.} \cref{eq:LoRAhubDistillation} assumes a wide range of pre-tuned LoRAs sampled from the underlying task distribution $p_{\mathcal{T}}$, which is crucial for enhancing generalization of meta-learning. However, a fixed number of LoRAs might be insufficient to capture the diversity of $p_{\mathcal{T}}$, especially within the setting of limited LoRA budgets. Therefore, we propose cross-task interpolation to densify the task distribution. We generate new tasks by combining classes from different pre-tuned LoRAs. For example, given LoRAs $\delta{W}_i$ and $\delta{W}_j$ tuned on tasks with classes (\textit{husky}, \textit{sparrow}) and (\textit{golden retriever}, \textit{wild horse}), an interpolated task might be (\textit{husky}, \textit{golden retriever}). This expands the range of tasks for meta-training, enhancing generalization. Since the interpolated task does not match the label spaces of any pre-tuned LoRAs, we modify \cref{eq:LoRAhubDistillation1} by replacing the KL loss with the Cross Entropy (CE) loss:
\begin{equation}
    \min_{\delta{W}^*} \mathbb{E}_{p_{\hat{\mathcal{T}}}
    } \sum_{(\mathbf{X}_{\rm q},\mathbf{Y}_{\rm q})\sim{\mathcal{D}}_{\rm q}^{\hat{\mathcal{T}}}} \text{CE}\left(P(\mathbf{Y}_{\rm pred} | \mathbf{X}_{\rm q}, {\mathcal{D}}_s^{\hat{\mathcal{T}}}; \delta{W}^*),\mathbf{Y}_{\rm q}\right),
\label{eq:interpolation}
\end{equation}
where $p_{\hat{\mathcal{T}}}$ refers to the interpolated task distribution and $(\mathcal{D}_{\rm s}^{\hat{\mathcal{T}}}, \mathcal{D}_{\rm q}^{\hat{\mathcal{T}}})$ refer to the generated support and query sets of the interpolated task $\hat{\mathcal{T}}$.


\subsection{Double-Efficient Mechanism}
\label{sec:double}
\textbf{Efficient inversion with token pruning.} As shown in \cref{eq:inversion}, optimizing $\mathbf{X}$ via LoRA Inversion requires iteratively forward and backward computations. To improve efficiency, we propose to prune unimportant tokens during inversion. This is reasonable since the self-attention mechanism inherently weights tokens according to their importance and relevance.
As illustrated in the left panel of \cref{fig:double}, at the $i^{\rm th}$ layer, we implement ``token pruning'' by directly discarding those unimportant tokens, no longer processing them forward or computing backward gradients, thus significantly reducing computational complexity.

\textit{The most important tokens are those with highest attention weights in} $\boldsymbol{a}_{\texttt{[CLS]}}$. Suppose we have $n+1$ tokens $[\boldsymbol{x}_{\texttt{[CLS]}}, \boldsymbol{x}_{1}, ..., \boldsymbol{x}_{n}]$ at the $i^{\rm th}$ layer,  where $\boldsymbol{x}_{\texttt{[CLS]}}$ is the class
token inserted before all image tokens to grasp global information. 
We propose to use the attention weights of the class token $\boldsymbol{x}_{\texttt{[CLS]}}$ with respect to all other tokens, as an indicator measuring each token's importance:
\begin{equation}
\boldsymbol{a}_{\texttt{[CLS]}}=\operatorname{Softmax}\left(\frac{\boldsymbol{q}_{\texttt{[CLS]}} \cdot {\boldsymbol{K}}^{\top}}{\sqrt{d}}\right),\label{eq:double1}
\end{equation}
where $\boldsymbol{a}_{\texttt{[CLS]}}$ is a $(n+1)$-dimension vector, representing the attention weights from token $\boldsymbol{x}_{\texttt{[CLS]}}$  to all tokens $[\boldsymbol{x}_{\texttt{[CLS]}},\boldsymbol{x}_1,...,\boldsymbol{x}_n]$.
$\boldsymbol{q}_{\texttt{[CLS]}}$ is the \textit{query vector} of token $\boldsymbol{x}_{\texttt{[CLS]}}$. $\boldsymbol{K}=[\boldsymbol{k}_{\texttt{[CLS]}}, \boldsymbol{k}_{1}, ..., \boldsymbol{k}_{n}]^{\top}$ is the \textit{key vectors} of all tokens. $d$ is the dimension of the \textit{query vector}.
The $\boldsymbol{a}_{\texttt{[CLS]}}$ is then used to calculate the output of token $\boldsymbol{x}_{\texttt{[CLS]}}$ via the self-attention mechanism:
\begin{equation}
\boldsymbol{x}_{\texttt{[CLS]}}=\boldsymbol{a}_{\texttt{[CLS]}}\cdot \boldsymbol{V},
\label{eq:double2}
\end{equation}
where $\boldsymbol{V}=[\boldsymbol{v}_{\texttt{[CLS]}}, \boldsymbol{v}_{1}, ..., \boldsymbol{v}_{n}]^{\top}$ is the \textit{value vectors} of all tokens. Therefore, the output of $\boldsymbol{x}_{\texttt{[CLS]}}$ can be viewed as a linear combination of all tokens' \textit{value vectors} weighted by $\boldsymbol{a}_{\texttt{[CLS]}}$. Since the output of $\boldsymbol{x}_{\texttt{[CLS]}}$ is used for classification at the final layer, it is rational to view $\boldsymbol{a}_{\texttt{[CLS]}}$ as an indicator, measuring the extent to which each token
contributes to final predictions, \textit{i.e.}, the importance of each token. Therefore, we identify the most important tokens as those with the highest attention weights in $\boldsymbol{a}_{\texttt{[CLS]}}$.
For multi-head self-attention, we compute average attention weights $\boldsymbol{a}_{\texttt{[CLS]}}$ across all heads. 
Note that this process requires no extra computational demands, as it is an inherent part of the forward process (see \cref{app:preliminary} for more preliminaries).

\noindent
\textbf{Efficient meta-training with sparse tokens.}\ We obtain the remaining tokens at the last layer from the inversion stage. Since each token (except for token $\boldsymbol{x}_{\texttt{[CLS]}}$) precisely corresponds to a token in the input image, these remaining tokens can indicate the most informative areas in the generated data, typically the foreground regions. 

\textit{Mask construction.} To highlight the most informative areas in the generated image,  we construct a mask matrix as shown in the right panel of \cref{fig:double}. The mask matrix is constructed by setting values of 1 at the positions of remaining tokens and 0 elsewhere. 

\textit{Selectively use unmasked tokens for meta-training.} We multiply the mask with the generated image to
create a masked image, reserving the most informative tokens (such as foregrounds) in the generated image. When meta-training the meta-LoRA, we only feed-forward the unmasked tokens. This selective use of sparse tokens significantly accelerates the meta-training process, while maintaining or even  improving performance by reducing
noise from the generated data (see \cref{tab:noise} in \cref{app:additionExperiments} for analysis).

\begin{algorithm}[t]
\DontPrintSemicolon
\small
\SetKwInOut{Input}{Input}\SetKwInOut{Output}{Output}\SetKwInOut{Require}{Require}
\textbf{INPUT} Then VFM $f$. Multiple pre-tuned LoRAs and classification heads. Coefficient $\alpha_{\mathcal{R}}$ in \cref{eq:inversion}.
\;
\textbf{OUTPUT} The meta-trained meta-LoRA $\delta{W}^*$\;
Randomly initialize the meta-LoRA $\delta{W}^*$\;
\While(){not done}{
\If {\textbf{\rm \textbf{not}} \text{\rm cross-task interpolation}}{
Randomly sample a LoRA and head $(\delta{W}, h_{\delta{W}})$\;
\tcp{Surrogate data generation}
Equip $f$ with $(\delta{W}, h_{\delta{W}})$\;
Generate surrogate data by minimizing \cref{eq:inversion}\;
(Optional) Transform into masked versions\;
Construct a meta-training task by splitting data to one support set and one query set $({\mathcal{D}}_{\rm s}^{{\mathcal{T}}}, {\mathcal{D}}_{\rm q}^{{\mathcal{T}}})$\;
\tcp{LoRA distillation}
Equip $f$ with the meta-LoRA $\delta{W}^*$\;
Make predictions on  ${\mathcal{D}}_{\rm q}^{{\mathcal{T}}}$ based on ${\mathcal{D}}_{\rm s}^{{\mathcal{T}}}$\ (\cref{eq:LoRAhubDistillation2})\;
Update $\delta{W}^*$ by minimizing \cref{eq:LoRAhubDistillation1}\;
}
\Else{
\tcp{Cross-task interpolation}
Construct the interpolated task $({\mathcal{D}}_s^{\hat{\mathcal{T}}}, {\mathcal{D}}_q^{\hat{\mathcal{T}}})$\;
Equip $f$ with the meta-LoRA $\delta{W}^*$\;
Make predictions on ${\mathcal{D}}_{\rm q}^{\hat{\mathcal{T}}}$ based on ${\mathcal{D}}_{\rm s}^{\hat{\mathcal{T}}}$\ (\cref{eq:LoRAhubDistillation2})\;
Update $\delta{W}^*$ by minimizing \cref{eq:interpolation}\;
}}
\caption{ LoRA Recycle}
\label{alg:training}
\end{algorithm}


%% file: sec/4_experiment.tex
\section{Experiments}
In this section, we perform comprehensive experiments on various few-shot classification benchmarks, covering both in-domain (see \cref{sec:indomain}) and cross-domain scenarios (see \cref{sec:crossdomain}). We also provide comprehensive visualization results and ablation studies in \cref{sec:ablation} and \cref{app:additionExperiments}.

\noindent
\textbf{Setup of VFM.} We select the 12-layer ViT-B/16 and ViT-B/32 pre-trained with CLIP as the pre-trained VFM, publicly available on HuggingFace. Refer to \cref{tab:vittypes} in \cref{app:additionExperiments} for more results on more types of transformer.


\noindent
\textbf{Baselines.}\ 
We compare LoRA Recycle against several baselines (see \cref{app:baselines} for more implementation details). 
\begin{itemize}[itemsep=5pt, parsep=-5pt,leftmargin=7pt]
    \item {Multi-LoRAs reuse baselines.} (a) {LoRAs Avg} averages all pre-tuned LoRAs into one, which is then either fine-tuned ({LoRAs Avg + Linear}) or used for Nearest Neighbor ({LoRAs Avg + NN}) inference. (b) {LoRAHub} \citep{huang2023LoRAhub} uses a weighted sum of pre-tuned LoRAs, with weights fine-tuned on the target task. (c) {MOLE} \citep{chen2024llava} fine-tunes a gating function to combine outputs from multiple LoRAs.
    \item {Fine-tuning-free baselines.} (d) {Nearest Neighbor (NN)} predicts based on the closest class center. (e) {CAML} \citep{fifty2023context} trains a sequence model to simulate in-context learning.
    \item {Few-shot learning with foundation models.} 
    (f) {$\text{P}>\text{M}>\text{F}$} \citep{hu2022pushing} is a state-of-the-art method that adapts foundation models to few-shot tasks through a pre-training, meta-training, and fine-tuning pipeline.
        \item {Fine-tuning baselines.} While our focus is on tuning-free settings, we include representative fine-tuning methods to demonstrate that tuning-free approaches can achieve comparable performance while offering advantages of stability and faster response. (g) {Full Fine-Tuning} updates the entire model, (h) {Linear Probe} updates only the classification head, and (i) {LoRA + Linear} \citep{hu2021LoRA} updates LoRA parameters alongside the classification head.
\end{itemize}


\begin{table*}[tbp]
  \centering
  \small
   \caption{ Recycle in-domain LoRAs. The VFM utilizes ViT-B/16 pre-trained by CLIP. \textbf{FT} refers to fine-tuning-based baselines and \textbf{FTF} refers to fine-tuning-free baselines. \textbf{LoRA Recycle$_x$} indicates using $x$\% token-masked images for meta-training. The superscripts represent performance gains over the best FT baselines, while the subscripts indicate gains over the best FTF baselines. 
   } 
   \vspace{-0.3cm}
   \scalebox{0.7}{
  \begin{tabular}{clcccccccc}
    \toprule
    \multirow{2}{*}{\textbf{ }}&
   \multirow{2}{*}{\textbf{Method}} 
    &\multicolumn{2}{c}{CIFAR-FS}&\multicolumn{2}{c}{MiniImageNet}&\multicolumn{2}{c}{VGG-Flower}&\multicolumn{2}{c}{CUB}\\
    \cmidrule(r){3-4}
    \cmidrule(r){5-6}
    \cmidrule(r){7-8}
    \cmidrule(r){9-10}
     & &\textbf{5-way 1-shot} &\textbf{5-way 5-shot} &\textbf{5-way 1-shot} &\textbf{5-way 5-shot}&\textbf{5-way 1-shot} &\textbf{5-way 5-shot}&\textbf{5-way 1-shot} &\textbf{5-way 5-shot} \\
    \midrule
    \multirow{6}{*}{\textbf{FT}}
    & Full Finetuning & 22.81  & 28.33 & 21.16  & 23.60 &  23.11  & 31.25 &  21.27  & 24.47 \\
    & Linear-probe &  80.06 &95.49  & 82.04  & 94.12 &  89.65  & 97.77 &  85.84  & 97.40  \\
    & LoRA + Linear &  79.29& 95.43 & 82.00  & 94.83 &  88.47  & 97.63 & 85.87  &  97.32 \\
    & $\text{P}>\text{M}>\text{F}$ &79.54&95.62&82.77&95.12&89.32&97.65&86.12&97.38\\
    & LoRAs Avg + Linear & 80.25  & 96.07 &  83.59 & 95.43 & 90.05   & 97.73 & 87.13   & 97.49  \\
    & MOLE&80.31&96.11&83.53&95.41&90.14&97.68&87.07&97.21\\
        &LoRAHub&  81.23 & 96.24 &  83.68 & 95.72 &  90.89  & 97.75 & 87.22   & 97.51  \\
    \midrule
    \multirow{8}{*}{\textbf{FTF}}
        & NN & 78.06  &94.09 & 81.08  &93.85  &  89.75  &97.78 &  85.11  &96.09   \\
    & LoRAs Avg + NN&  79.37 & 93.45 & 81.72  & 94.64 &  90.08  & 97.92 &  85.16  & 97.23  \\
       & CMAL&  81.02 & 93.59& 81.89  & 94.81 & 91.10   & 97.98 &  86.51  &  97.32 \\
    \cmidrule(r){2-10}
    & \textbf{LoRA Recycle } & 89.69&\textbf{97.05 }$_{\textcolor{blue}{(+2.96\%)}}^{\textcolor{blue}{(+0.81\%)}}$&\textbf{88.60  }$_{\textcolor{blue}{(+6.71\%)}}^{\textcolor{blue}{(+4.92\%)}}$&96.12  & \textbf{94.53}$_{\textcolor{blue}{(+3.43\%)}}^{\textcolor{blue}{(+3.64\%)}}$  & 98.59   &  91.12  & 97.67  \\
    & \textbf{LoRA Recycle$_{\textbf{25}}$ } & \textbf{91.03}$_{\textcolor{blue}{(+10.01\%)}}^{\textcolor{blue}{(+9.80\%)}}$ & 96.53 &  87.51 & \textbf{96.25}$_{\textcolor{blue}{(+1.41\%)}}^{\textcolor{blue}{(+0.53\%)}}$ & 94.38 & 98.53 &  90.16  &  97.48   \\
    & \textbf{LoRA Recycle$_{\textbf{50}}$ } & 90.91 & 96.08 & 87.21  & 95.85 &  94.05  & 98.56 &   90.65 &  97.41 \\
    & \textbf{LoRA Recycle$_{\textbf{75}}$ } & 89.70  &  96.69&  87.36 & 96.05 &  94.28  & \textbf{98.76}$_{\textcolor{blue}{(+0.78\%)}}^{\textcolor{blue}{(+0.99\%)}}$ &  \textbf{91.21}$_{\textcolor{blue}{(+4.70\%)}}^{\textcolor{blue}{(+3.99\%)}}$  &  \textbf{98.23}$_{\textcolor{blue}{(+0.91\%)}}^{\textcolor{blue}{(+0.72\%)}}$ \\
     \bottomrule
     \end{tabular}}
  \label{tab:single16}
    \vspace{-0.4cm}
\end{table*}
\begin{table*}[tbp]
  \centering
   \caption{ Recycle cross-domain LoRAs.  The VFM utilizes ViT-B/16 pre-trained by CLIP. 
   } 
   \vspace{-0.3cm}
   \scalebox{0.66}{
  \begin{tabular}{clcccccccc}
    \toprule
    \multirow{2}{*}{\textbf{ }}&
   \multirow{2}{*}{\textbf{Method}} 
    &\multicolumn{2}{c}{ChestX}&\multicolumn{2}{c}{ISIC}&\multicolumn{2}{c}{EuroSAT}&\multicolumn{2}{c}{CropDiseases}\\
    \cmidrule(r){3-4}
    \cmidrule(r){5-6}
    \cmidrule(r){7-8}
    \cmidrule(r){9-10}
     & &\textbf{5-way 1-shot} &\textbf{5-way 5-shot} &\textbf{5-way 1-shot} &\textbf{5-way 5-shot}&\textbf{5-way 1-shot} &\textbf{5-way 5-shot}&\textbf{5-way 1-shot} &\textbf{5-way 5-shot} \\
    \midrule
    \multirow{6}{*}{\textbf{FT}}
    & Full Finetuning & 20.12  & 20.00 & 21.33  & 26.21 &  25.55  & 36.11 &  22.45  & 28.48  \\
    & Linear-probe &  21.20 & 24.00 &  31.17 & 43.60 &   62.64 & 83.91 &  77.48  &  92.57 \\
    & LoRA + Linear &  21.05 & 22.37 &  30.72 & 45.16 &   68.13 & 88.68 &  77.33  &  94.19 \\
    &$\text{P}>\text{M}>\text{F}$ &21.12   &22.21  &30.77   &45.54  &68.51    &88.71  &77.65    &94.21   \\
    & LoRAs Avg + Linear &  21.37 & 20.84 & 30.51  & 45.88 & 68.77   & 88.29 & 78.75   &  94.37 \\
    &MOLE & 21.24  &20.67  &30.61   &45.79  &68.84    &88.42  &78.81    &94.40   \\
    &LoRAHub  & 21.45  & 22.61 & 32.11  & 46.12 & 69.45   & \textbf{89.76} &  79.32  &  94.44 \\
    \midrule
    \multirow{8}{*}{\textbf{FTF}}
        & NN &  21.23 & 22.84 & 31.20  & 40.58 & 61.73   & 80.05 &   75.48 &  91.89 \\
    & LoRAs Avg + NN& 20.80  & 23.04 & 29.67  & 39.56 &  62.52  & 78.87 &  78.91  & 91.57 \\
    &CMAL  &  21.26 & 23.24 & 29.97  & 41.27 &  67.69  & 83.87 &  79.71  & 93.38  \\
    \cmidrule(r){2-10}
    & \textbf{LoRA Recycle } &  22.32 & 24.61 & 33.76 & 47.96 &  66.95  & 85.17 & 83.07  & 95.33 \\
    & \textbf{LoRA Recycle$_{\textbf{25}}$ } & 22.77 & 24.88 &  33.64 & \textbf{48.29}$_{\textcolor{blue}{(+7.02\%)}}^{\textcolor{blue}{(+2.41\%)}}$ &  67.65  & 84.73 &  82.41  & 95.40  \\
    & \textbf{LoRA Recycle$_{\textbf{50}}$ } & \textbf{23.08}$_{\textcolor{blue}{(+1.82\%)}}^{\textcolor{blue}{(+1.63\%)}}$ &  \textbf{25.43}$_{\textcolor{blue}{(+2.19\%)}}^{\textcolor{blue}{(+1.43\%)}}$& \textbf{35.31}$_{\textcolor{blue}{(+4.11\%)}}^{\textcolor{blue}{(+3.20\%)}}$  & 47.41 &  \textbf{69.88}$_{\textcolor{blue}{(+2.19\%)}}^{\textcolor{blue}{(+0.43\%)}}$  & 86.72 & \textbf{83.63}$_{\textcolor{blue}{(+3.92\%)}}^{\textcolor{blue}{(+4.31\%)}}$   & \textbf{96.33}$_{\textcolor{blue}{(+2.95\%)}}^{\textcolor{blue}{(+1.89\%)}}$  \\
    & \textbf{LoRA Recycle$_{\textbf{75}}$ } & 22.99 & 24.91 & 35.16  & 48.25 &  68.00  & 87.98$_{\textcolor{blue}{(+4.11\%)}}$ &  81.92  &  95.64 \\
     \bottomrule
  \end{tabular}}
  \label{tab:cross16}
    \vspace{-0.3cm}
\end{table*}

\noindent
\textbf{Implementation details.} We fine-tune the LoRAs (rank \( r = 4 \)) and classification heads using the Adam optimizer with a learning rate of \( 1 \times 10^{-3} \). LoRA performs well with sufficient data. During meta-training, the meta-LoRA is optimized with Adam at the same learning rate, following a cyclic schedule: a 25-iteration linear warm-up from \( 1 \times 10^{-5} \) to \( 1 \times 10^{-3} \), followed by cosine annealing over the next 75 iterations. For LoRA Inversion, surrogate data is optimized using Adam with a learning rate of 0.25 over 2000 iterations, and generated images have a resolution of \( 224 \times 224 \). We set the hyperparameter \( \alpha_{\mathcal{R}} = 0.01 \). Data augmentation includes random horizontal flipping and normalization in meta-training, with only normalization applied in meta-testing. Hyperparameter selections and sensitivity analysis are discussed in \cref{app:hyperparameter}. Unless otherwise specified, we perform token pruning in the final layer for the inversion stage to obtain masks with varying sparsity.

\subsection{Recycle In-Domain LoRAs}
\label{sec:indomain}
\textbf{In-domain benchmarks.} For the ``recycle in-domain LoRAs" scenario, we use four datasets commonly used to evaluate few-shot adaptability: CIFAR-FS \citep{bertinetto2018meta}, MiniImageNet \citep{vinyals2016matching}, VGG-Flower \citep{flower}, and CUB \citep{cub}. These datasets range from general natural images (CIFAR-FS, MiniImageNet) to more specialized domains, such as bird species (CUB) and flowers (VGG-Flower). Following standard meta-learning splits \citep{finn2017model}, each dataset is divided into meta-training and meta-testing subsets with non-overlapping label spaces. For constructing an $N$-way $K$-shot task, we randomly sample $N$ classes and $K$ examples per class from the corresponding subset.

\noindent
\textbf{In-domain setup.} We collect 100 LoRAs pre-tuned on diverse 5-way tasks constructed from one specific meta-training subset. Our evaluation, in contrast, is based on meta-testing tasks constructed from  the corresponding meta-testing subset. This setup ensures the pre-tuned LoRAs and meta-testing tasks originate from the same domain (dataset) but with non-overlapping label spaces.

\noindent
\textbf{Results in in-domain scenario.} \cref{tab:single16} shows the results for the ``recycle in-domain LoRAs'' scenario. Notable findings are as follows: ``LoRA Recycle'' surpasses the best fine-tuning-based baselines by considerable margins, especially up to 9.80\% for 1-shot learning. It also outperforms the top fine-tuning-free baselines by up to 10.01\% for 1-shot learning, confirming its superior adaptability without the need for fine-tuning.  ``Full fine-tuning"  performs the worst, as it tends to overfit when tuning large models with extremely few examples. This issue also leads to poor performance of ``$\text{P}>\text{M}>\text{F}$", though the second meta-training stage helps mitigate it to some extent.
``LoRAs Avg'' and ``LoRAHub'' do not ensure effective generalization to new tasks. The reason is that each pre-tuned LoRA targets different tasks, and the arithmetic operation like averaging in the parameter space lacks precise alignment among different LoRAs. 
Moreover, these baselines do not explicitly incorporate meta-learning objectives, which have proven to be useful for enhancing generalization in few-shot learning. More discussions on the different performance gains across datasets are provided in \cref{app:discussion}.

\noindent
\textbf{Unique advantage of token pruning on generated data beyond acceleration.} Meta-training with sparse tokens can bring performance gains up to 1.34\%. This is a unique advantage of token pruning on generated data, as the inversion process typically crafts only label-relevant features (\ie, the foreground) into the generated data, while other regions (\ie, the background) often remain noisy as initialization (see \cref{tab:noise} in \cref{app:additionExperiments} for evidence). Pruning background tokens helps reduce noise from the generated data.

\begin{figure*}[!t]
  \centering
    \includegraphics[width=0.9\linewidth]{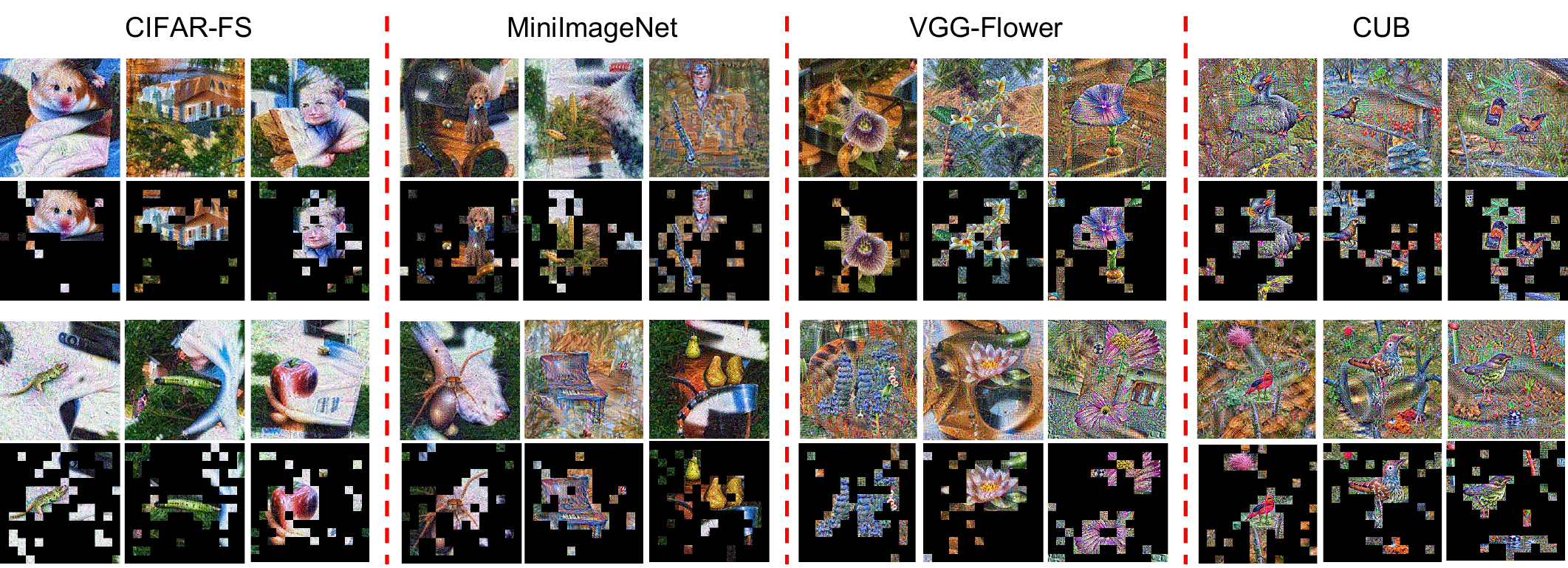}
   \vspace{-0.2cm}
   \caption{Visualization of generated images with their 75\% token-masked versions.}
   \label{fig:vis}
   \vspace{-0.4cm}
\end{figure*}

\subsection{Recycle Cross-Domain LoRAs}
\label{sec:crossdomain}
\textbf{Cross-domain benchmarks.} Real-world situations might pose challenges in collecting LoRAs from the identical domain. For the ``recycle cross-domain LoRAs" scenario, we construct meta-training tasks from the meta-training subsets of CIFAR-FS, MiniImageNet, VGG-Flower, and CUB, while meta-testing tasks are drawn from distinct datasets (ChestX, ISIC, EuroSAT, or CropDiseases) following the cross-domain meta-learning benchmark \citep{guo2020broader}. These meta-testing datasets cover diverse domains, including medical images (ChestX, ISIC) and specialized imagery (EuroSAT, CropDiseases), ensuring a distinct domain difference between meta-training and meta-testing. Results on Meta-Dataset \cite{metadataset} is provided in \cref{tab:metadataset} in \cref{app:additionExperiments}.

\noindent
\textbf{Cross-domain setup.} We collect 100 LoRAs pre-tuned on diverse 5-way tasks constructed from four meta-training subsets, including CIFAR-FS, MiniImageNet, VGG-Flower and CUB. Our evaluation, in contrast, is based on meta-testing tasks from one specific cross-domain dataset (ChestX, ISIC, EuroSAT or CropDiseases). 
This ensures the pre-tuned LoRAs and meta-testing tasks originate from distinctly different domains (datasets) and also with strictly non-overlapping label spaces. 

\noindent
\textbf{Results in cross-domain scenario.} \cref{tab:cross16} shows the results for the challenging “recycle cross-domain LoRAs''. Existing methods struggle under these conditions, reflecting the inherent difficulty of this scenario. Despite this, LoRA Recycle achieves notable improvements, outperforming the best fine-tuning-based baselines by up to 4.31\% and 2.41\% for 1-shot and 5-shot learning, respectively. It also exceeds the top tuning-free baselines by up to 4.11\% and 7.02\% for 1-shot and 5-shot learning, confirming its enhanced cross-domain robustness. However, while LoRA Recycle greatly advances  performance, there remains room for improvement when facing substantial distribution shifts, offering a promising direction for future research.

\subsection{Ablation Studies}
\label{sec:ablation}

\textbf{How to choose the overall pruning ratio and pruning layers during inversion?} We adopt the double-efficient mechanism with two steps: (i) Set the overall pruning ratio (sparsity level) of the generated data. As shown in \cref{tab:single16} and \cref{tab:cross16}, pruning 50\% or 75\% of tokens boosts performance while preserving key foregrounds (see \cref{fig:vis}). (ii) Select the pruning layers during inversion. Pruning at different layers affects only the inversion speed, while meta-training speed depends solely on the overall pruning ratio. The choice of pruning layers is flexible and depends on needs: with the same overall pruning ratio, \cref{tab:strategy} suggests pruning at deeper layers for better performance, or shallower layers for faster inversion. The choice of middle-layer pruning or multi-layer pruning across shallow and deep layers can balance the trade-off between efficiency and performance to some extent.

\noindent
\begin{table}[t]
\centering
\caption{Complexity analysis of inversion. \{\textbf{x}: y\} denotes pruning (y $\times$ 100)\% tokens at the $\mathbf{x}^{\rm th}$ layer. Measurements are recorded during inversion with a batch size of 25 on CIFAR-FS.} 
\vspace{-0.3cm}
  \scalebox{0.58}{\begin{tabular}{ccccccc}
    \toprule
     \multirow{2}{*}{\shortstack{\textbf{Token Pruning}\\\textbf{Strategy}}} & \multirow{2}{*}{\textbf{5w 1s}}&\multirow{2}{*}{\textbf{5w 5s}}& \multirow{2}{*}{\shortstack{\textbf{Throughput} \\({\rm its/s})  \color{blue}$\uparrow$}} & \multirow{2}{*}{\shortstack{\textbf{FLOPs} \\({\rm G}) \color{blue}$\downarrow$}}& \multirow{2}{*}{\shortstack{\textbf{GPU Mem} \\({\rm GB}) \color{blue}$\downarrow$}}\\
     & & & & &\\
    \midrule
    \{\textbf{0}: 0.0\} &  89.69  & 97.05  & 5.56 &50.59& 8.74\\
    \midrule
     \{\textbf{11}: 0.75\}  &  89.43  & 96.72 & 5.81 \color{blue}(+4\%)&48.51 \color{blue}(-4\%)&8.63 \color{blue}(-1\%)\\
      \{\textbf{8}: 0.75\} & 82.27   & 95.69  & 6.22 \color{blue}(+12\%)&39.14 \color{blue}(-23\%)&8.07 \color{blue}(-8\%)\\
     \{\textbf{6}: 0.75\}  &  81.08  & 95.52 & 7.15 \color{blue}(+29\%)&32.89 \color{blue}(-35\%)& 7.69 \color{blue}(-12\%)\\   
     \midrule
     \{\textbf{3}: 0.3, \textbf{6}: 0.3, \textbf{8}: 0.3, \textbf{11}: 0.3\}  &84.17  & 96.12 & 6.13 \color{blue}(+10\%)&40.00 \color{blue}(-21\%)&8.08 \color{blue}(-8\%)\\
     \bottomrule
  \end{tabular}}
  \label{tab:strategy}
\end{table}

\noindent
\textbf{Effect of sparse ratio of the generated data for the performance and complexity of meta-training.}\ As shown in \cref{tab:mask}, discarding 75\% of tokens in the generated data significantly accelerates meta-training by up to $3\times$ and yields performance gains of up to +0.56\% on CUB by reducing noise from the generated data.

\begin{table}[t]
\vspace{-0.cm}
\centering
\caption{Effect of sparse ratio of the generated data for the performance and complexity of meta-training. Measurements are recorded during meta-training with a batch size of 100 on CUB.} 
\vspace{-0.3cm}
  \scalebox{0.66}{\begin{tabular}{lccccc}
    \toprule
     \multirow{2}{*}{\textbf{Sparse ratio}} & \multirow{2}{*}{\textbf{5w 1s}}&\multirow{2}{*}{\textbf{5w 5s}}& \multirow{2}{*}{\shortstack{\textbf{Throughput}\\ ({\rm its/s}) \color{blue}$\uparrow$}}& \multirow{2}{*}{\shortstack{\textbf{FLOPs} \\({\rm G}) \color{blue}$\downarrow$}}& \multirow{2}{*}{\shortstack{\textbf{GPU Mem} \\({\rm GB}) \color{blue}$\downarrow$}}\\
     &&&&&\\
    \midrule
    {LoRA Recycle} & 91.12 & 97.67 &1.76& 50.59&12.86\\ 
    \midrule
    {LoRA Recycle$_{{25}}$} & 90.16 & 97.48&2.34 \color{blue}(+33\%) &38.09 \color{blue}(-25\%)& 9.40 \color{blue}(-27\%)\\ 
    {LoRA Recycle$_{{50}}$} & 90.65 & 97.41&3.63 \color{blue}(+106\%)& 25.60 \color{blue}(-49\%)& 6.23 \color{blue}(-52\%)\\ 
    {LoRA Recycle$_{{75}}$} & 91.21 & 98.23&6.83 \color{blue}(+287\%)& 13.10 \color{blue}(-74\%)&3.31 \color{blue}(-74\%)\\ 
     \bottomrule
  \end{tabular}}
  \label{tab:mask}
  \vspace{-0.2cm}
\end{table}

\noindent
\textbf{Visualization.}\ As shown in \cref{fig:vis}, the generated data effectively retains semantic foregrounds while filtering out noisy backgrounds, resulting in high-resolution 224 × 224 images better than existing methods (see \cref{fig:vis_com} in \cref{app:additionExperiments}).

\noindent
\textbf{Comprehensive ablation studies} are provided in \cref{app:additionExperiments} due to limited space, to thoroughly demonstrate the effect and design rationale of each component in our framework.

%% file: sec/sup.tex
\clearpage
\vspace{\baselineskip}
\begin{center}
\LARGE
\textbf{Appendix}
\end{center} 
\begin{appendix}

\section{Additional Experiments}
\label{app:additionExperiments}


\noindent
\textbf{Advantage of token pruning in generated data beyond acceleration.} When using inversion to generate data by optimizing \cref{eq:inversion}, it primarily crafts only the label-relevant features  into the generated data (typically the foreground regions), while background regions often remain noisy as initialization. To further illustrate this, the experiment in \cref{tab:noise} shows that, during the inversion process, the classification loss \cref{eq:inversion} caused by the identified foreground steadily decreases, whereas the  loss caused by the identified background remains nearly unchanged. As such, masking the background in generated data helps to reduce noise, which aligns with our experimental results in \cref{tab:single16}: pruning background tokens in the generated CIFAR-FS data led to a 1.34\% improvement in performance.

\begin{table}[!h]
\centering
\caption{Loss tracking during inversion process.} 
\vspace{-0.3cm}
  \scalebox{0.7}{\begin{tabular}{cc}
    \toprule
     \textbf{Area} & \textbf{Classification Loss Change in \cref{eq:inversion}}\\
     \midrule
      Inverted Backgrounds&10.78 $\rightarrow$ 10.72\\
     Inverted Foregrounds&10.78 $\rightarrow$ 0.12\\
     \bottomrule
  \end{tabular}}
  \label{tab:noise}
\end{table}

\noindent
\textbf{Effect of cross-task interpolation.}\ \cref{tab:interpolation} verifies the effectiveness of the cross-task interpolation under a constrained LoRA budget of 100 on CIFAR-FS. This technique can diversify the task distribution by generating multiple interpolated tasks, which enables the meta-training to cover a broader range of tasks, thereby bolstering the generalization capabilities for unseen tasks.

\begin{table}[!h]
\centering
\caption{Effect of cross-task interpolation.} 
\vspace{-0.3cm}
  \scalebox{0.7}{\begin{tabular}{ccc}
    \toprule
     \textbf{Ablation} & \textbf{5-way 1-shot} & \textbf{5-way 5-shot} \\
     \midrule
      \textbf{w/o} cross-task interpolation&87.97&96.81\\
     \textbf{w/} cross-task interpolation&89.69&97.05\\
     \bottomrule
  \end{tabular}}
  \label{tab:interpolation}
\end{table}

\noindent
\textbf{Effect of meta-learning in LoRA Recycle.}\ To assess the effectiveness of meta-learning, we compared it against joint supervised learning within the LoRA Recycle framework. In joint supervised learning, data generated from all LoRAs is aggregated to train a single LoRA through standard supervised learning. The comparative results for meta-learning and joint supervised learning are presented in \cref{tab:meta-learning}.  Experiments were conducted on the CIFAR-FS dataset, focusing on unseen few-shot tasks to evaluate generalization capabilities. As shown, meta-learning achieves significantly better performance than joint supervised learning in both 1-shot and 5-shot settings. This improvement arises because meta-learning’s bi-level optimization is inherently designed to enhance generalization to unseen few-shot tasks.

\begin{table}[!h]
\centering
\caption{Effect of meta-learning in LoRA Recycle.} 
\vspace{-0.3cm}
  \scalebox{0.7}{\begin{tabular}{ccc}
    \toprule
     \textbf{Ablation} & \textbf{5-way 1-shot} & \textbf{5-way 5-shot} \\
     \midrule
      joint supervised learning &78.22&94.56\\
     meta-learning &89.69&97.05\\
     \bottomrule
  \end{tabular}}
  \label{tab:meta-learning}
\end{table}

\noindent
\textbf{Effect of generated data in LoRA Recycle.}\ In our setting, the original training data for each LoRA is unavailable. To evaluate the effectiveness of the generated data, we use the results obtained from the original training data as an upper bound for comparison. As shown in Table \ref{tab:generated_data}, the performance on CIFAR-FS achieved with generated data is close to that obtained with the original training data, demonstrating the effectiveness of the generated data in LoRA Recycle.

\begin{table}[!h]
\centering
\caption{Effect of generated data in LoRA Recycle.} 
\vspace{-0.3cm}
  \scalebox{0.7}{\begin{tabular}{ccc}
    \toprule
     \textbf{Ablation} & \textbf{5-way 1-shot} & \textbf{5-way 5-shot} \\
     \midrule
      original training data&91.21&98.93\\
     generated data&89.69&97.05\\
     \bottomrule
  \end{tabular}}
  \label{tab:generated_data}
\end{table}

\begin{figure*}[!h]
  \centering
\includegraphics[width=0.6\linewidth]{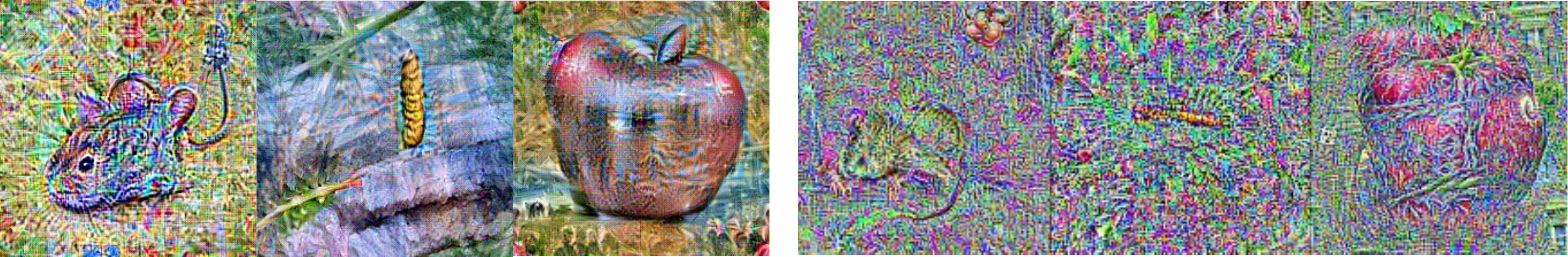}
   \vspace{-0.3cm}
   \caption{Visualization of generated images with (left) and without (right) the naturalness prior $\mathcal{R}_{\rm BN}$.}
   \label{fig:bn}
   \vspace{-0.2cm}
\end{figure*}

\noindent
\textbf{Meta-learn what?}\ Our framework meta-trains an extra lightweight LoRA while keeping the original VFM frozen. Based on the results shown in \cref{tab:variant}, we summarize some findings: (i) Meta-training the entire VFM is inferior to only meta-training the extra LoRA. Meta-training the entire VFM might distort the original feature space \citep{kumar2022fine}, leading to bias to meta-training tasks and heavy costs of computation and storage. Meta-training the extra LoRAs can preserve the knowledge of foundation models learned from large-scale pretraining while injecting task-specific knowledge into extra LoRAs. (ii) Only meta-training the last  6   LoRA layers can  outperform meta-training all LoRA layers. The improvements are more obvious in 5-way 1-shot learning, suggesting that reducing learnable parameters possibly avoids overfitting with limited training data. Only meta-training the first 6  LoRA layers is less effective. This is because only updating the shallow layers is insufficient to develop effective representations compared with updating the deep layers.

\begin{table}[!h]
  \centering
   \caption{Meta-learn what?} 
   \vspace{-0.3cm}
  \scalebox{0.8}{\begin{tabular}{ccccccc}
    \toprule
     \multirow{2}{*}{\textbf{Learnable Parts}} & \multirow{2}{*}{\textbf{5-way 1-shot}}&\multirow{2}{*}{\textbf{5-way 5-shot}}\\
     & & \\
    \midrule
    Entire VFM& 88.40& 95.73\\
    LoRA (all 12 layers)& 89.69& \textbf{97.05}\\
          LoRA (the first 6 layers)&85.45&95.13\\
          LoRA (the last 6 layers)&\textbf{90.40}&96.10\\
     \bottomrule
  \end{tabular}}
  \label{tab:variant}
\end{table}

\noindent
\textbf{Effect of constructed mask.}\ We use the constructed mask instead of attention weights to prune tokens before the first layer in meta-training. This choice is motivated by the observation that shallow-layer attention weights are less accurate than deep-layer weights  \cite{rao2021dynamicvit,kim2024token}, as shallow layers lack the fine-grained information captured by deeper layers. By leveraging final-layer attention information, our constructed mask offers more accurate guidance to identify the most informative tokens in the generated image. As shown in \cref{tab:mask_attention}, under the same pruning ratio (75\%), using the constructed mask yields significantly better performance than using first-layer attention on CIFAR-FS.

\begin{table}[!h]
\centering
\caption{Effect of constructed mask.} 
\vspace{-0.3cm}
  \scalebox{0.7}{\begin{tabular}{ccc}
    \toprule
     \textbf{Ablation} & \textbf{5-way 1-shot} & \textbf{5-way 5-shot} \\
     \midrule
      first-layer attention weight&83.62&92.43\\
     constructed mask&89.70&96.69\\
     \bottomrule
  \end{tabular}}
  \label{tab:mask_attention}
\end{table}

\noindent
\textbf{Experiments on more challenging dataset, Meta-Dataset} \citep{metadataset}. We evaluate our proposed LoRA Recycle framework on the Meta-Dataset, a benchmark specifically designed to test few-shot learning models across a variety of challenging domains. This dataset provides a rigorous evaluation setting.
The results, summarized in \cref{tab:metadataset}, demonstrate the effectiveness of LoRA Recycle compared to other baselines. Notably, LoRA Recycle achieves superior performance in both the 5-way 1-shot and 5-way 5-shot settings, while also offering the advantage of being fine-tuning-free.

\begin{table}[!h]
  \centering
   \caption{Experiments on more challenging Meta-Dataset.} 
   \vspace{-0.3cm}
  \scalebox{0.7}{\begin{tabular}{cccc}
    \toprule
     \textbf{Method}&\textbf{Fine-Tuning-Free} & \textbf{5-way 1-shot} & \textbf{5-way 5-shot} \\
     \midrule
     MOLE&\XSolidBrush&61.87&76.31\\
     LoRAHub&\XSolidBrush&63.14&77.24\\
     LoRA Recycle (ours)&\Checkmark&68.48&80.12\\
     \bottomrule
  \end{tabular}}
  \label{tab:metadataset}
\end{table}

\noindent
\textbf{Experiments on more types of Vision Transformers.} In this section, we evaluate the performance of our LoRA Recycle framework across multiple Vision Transformers on the CIFAR-FS dataset, further demonstrating its generalizability. We experiment with three popular Vision Transformer architectures: ViT-B (CLIP), DeiT-B \citep{touvron2021training}, and LV-ViT-M \citep{jiang2021all}. Each model is compared using LoRAHub as a baseline. \cref{tab:vittypes} presents the results of these experiments. The performance is evaluated in both 5-way 1-shot and 5-way 5-shot scenarios. As shown, LoRA Recycle consistently outperforms the LoRAHub baseline, while also offering the advantage of being fine-tuning-free.

\begin{table}[h]
  \centering
   \caption{Experiments on more types of Vision Transformers on CIFAR-FS.}
   \vspace{-0.3cm}
  \scalebox{0.6}{\begin{tabular}{ccccc}
    \toprule
     \textbf{Model}&\textbf{Method}&\textbf{Fine-Tuning-Free} & \textbf{5-way 1-shot} & \textbf{5-way 5-shot} \\
     \midrule
     \multirow{2}{*}{ViT-B (CLIP)}&LoRAHub&\XSolidBrush&81.02&96.24\\
     &LoRA Recycle (ours)&\Checkmark&91.03&97.05\\
     \midrule
     \multirow{2}{*}{DeiT-B \citep{touvron2021training}}&LoRAHub&\XSolidBrush&79.52&93.32\\
     &LoRA Recycle (ours)&\Checkmark&88.31&94.72\\
     \midrule
     \multirow{2}{*}{LV-ViT-M \citep{jiang2021all}}&LoRAHub&\XSolidBrush&80.42&94.23\\
     &LoRA Recycle (ours)&\Checkmark&89.52&95.35\\
     \bottomrule
  \end{tabular}}
  \label{tab:vittypes}
\end{table}

\noindent
\textbf{Experiments on tasks beyond few-shot learning.} In this section, we extend our evaluation to zero-shot classification tasks, demonstrating the versatility of LoRA Recycle beyond few-shot learning. To enable zero-shot classification, we recycle pre-tuned LoRAs from CLIP by replacing the classification loss used in \cref{eq:inversion} and \cref{eq:interpolation} with the contrastive loss employed by CLIP.
\cref{tab:zeroshot} presents the results on the Meta-Dataset for zero-shot classification. As shown, LoRA Recycle significantly outperforms other baseline methods, including MOLE and LoRAHub, both of which require fine-tuning. LoRA Recycle, being fine-tuning-free, achieves a higher accuracy, illustrating its effectiveness in adapting to zero-shot classification tasks.

\begin{table}[h]
  \centering
   \caption{Experiments on zero-shot classification on Meta-Dataset.} 
   \vspace{-0.3cm}
  \scalebox{0.8}{
  \begin{tabular}{cccc}
    \toprule
     \textbf{Method}&\textbf{Fine-Tuning-Free} & \textbf{5-way 0-shot} \\
     \midrule
     MOLE&\XSolidBrush&59.36\\
     LoRAHub&\XSolidBrush&60.25\\
     LoRA Recycle (ours)&\Checkmark&64.52\\
     \bottomrule
  \end{tabular}
  }
  \label{tab:zeroshot}
\end{table}

\noindent
\textbf{Recycle LoRAs with different ranks.}\ \cref{tab:rank} verifies the architecture-agnostic feature of our LoRA Recycle approach. Our approach can reuse pre-tuned LoRAs with different ranks (e.g., 50\% LoRAs with the rank of 4 and 50\% LoRAs with the rank of 8). This is a distinctive advantage absent in existing baselines, thereby extending its practical  applicability across various real-world scenarios.
\begin{table}[h]
  \centering
   \caption{Architecture-agnostic property of our framework. We conduct experiments on CIFAR-FS and set the rank of meta-LoRA as 4. We reuse pre-tuned LoRAs with different ranks (e.g., 50\% LoRAs with the rank of 4 and 50\% LoRAs with the rank of 8).} 
   \vspace{-0.3cm}
  \scalebox{0.8}{\begin{tabular}{ccc}
    \toprule
     \textbf{Rank of pre-tuned LoRAs} & \textbf{5-way 1-shot} & \textbf{5-way 5-shot} \\
     \midrule
     100\%: \textbf{4}&89.69&97.05\\
     50\%: \textbf{4} + 50\%: \textbf{8}&90.67&97.12\\
     \bottomrule
  \end{tabular}}
  \label{tab:rank}
\end{table}

\noindent
\textbf{Cross validation.}\ \cref{tab:cross_validation} shows our consistent superiority compared with other baselines by exchanging meta-training and meta-testing domains.

\begin{table}[!h]
    \centering
    \caption{Cross validation by exchanging meta-training and meta-testing domains. [meta-training domains]$\rightarrow$[meta-testing domain]. $\mathcal{D}_1$: MiniImageNet, $\mathcal{D}_2$: CUB, $\mathcal{D}_3$: CropDiseases. 51: 5-way 1-shot. 55: 5-way 5-shot.}
    \vspace{-0.3cm}
    \scalebox{0.6}{
    \begin{tabular}{ccccccc}
    \toprule
         \multirow{2}{*}{\textbf{Method}}&\multicolumn{2}{c}{$[D_2, D_3]\rightarrow[D_1]$}&\multicolumn{2}{c}{$[D_1, D_3]\rightarrow[D_2]$}&\multicolumn{2}{c}{$[D_1, D_2]\rightarrow[D_3]$}\\
         \cmidrule(r){2-3}
         \cmidrule(r){4-5}
         \cmidrule(r){6-7}
         &51&55&51&55&51&55\\
    \midrule
    LoRAHub + NN &81.02&93.18&85.27&95.23&76.21&92.31\\
    \midrule
    LoRA Recycle$_{75}$ (ours)&\textbf{86.12}&\textbf{95.03}&\textbf{90.02}&\textbf{97.12}&\textbf{80.19}&\textbf{94.02}\\
    \bottomrule
    \end{tabular}
    }    
    \label{tab:cross_validation}
\end{table}

\begin{figure*}[htbp]
  \centering
    \includegraphics[width=0.95\linewidth]{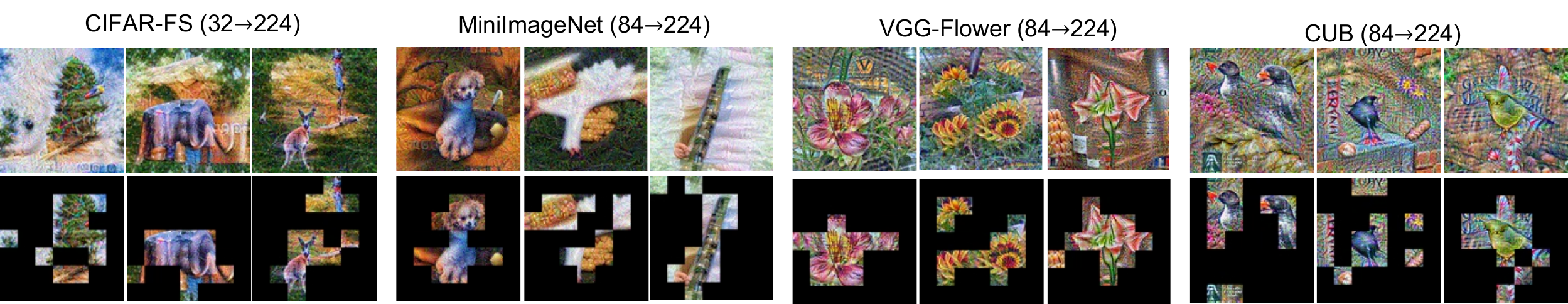}
   \caption{Visualization of generated images (odd line) and their 75\% token-masked versions (even line) from ViT-B/32. (32 $\rightarrow$ 224) denotes the original training images' resolution is 32 $\times$ 32 while we can reconstruct images with a higher resolution of 224 $\times$ 224. Note that the size of each patch is 32 $\times$ 32, instead of 16 $\times$ 16.}
   \label{fig:vis32}
   \vspace{-0.1cm}
\end{figure*}

\noindent
\textbf{Results of ViT-B/32.}\ \cref{tab:single32}
show the results when using ViT-B/32 with a 32 $\times$ 32 input patch size as the implementation of VFM. In the ``recycle in-domain LoRAs'' scenario, our LoRA Recycle consistently outperforms the best fine-tuning-based baselines by a large margin, up to 8.93\% and 1.40\% for 1-shot and 5-shot learning, respectively. It also exceeds the leading fine-tuning-free baselines by up to 10.39\% and 2.89\% for 1-shot and 5-shot learning, respectively. 
\cref{fig:vis32} shows the visualization of generated images and their masked versions generated from  ViT-B/32.

\begin{table*}[t]
  \centering
   \caption{ Recycle in-domain LoRAs. VFM is implemented with ViT-B/32. \textbf{FT} refers to fine-tuning-based baselines and \textbf{FTF} refers to fine-tuning-free baselines. \textbf{LoRA Recycle$_x$} indicates $x$\% tokens in generated data are masked (\textit{i.e.}, different sparsity ratios). For a fair comparison between different sparsity ratios, we perform token pruning at the same layer (\textit{i.e.}, at the last layer). Superscripts represent performance gains over the best FT baselines, while subscripts indicate gains over the best FTF baselines.} 
   \vspace{-0.3cm}
   \scalebox{0.57}{
  \begin{tabular}{clcccccccc}
    \toprule
    \multirow{2}{*}{}&
   \multirow{2}{*}{\textbf{Method}} 
    &\multicolumn{2}{c}{CIFAR-FS}&\multicolumn{2}{c}{MiniImageNet}&\multicolumn{2}{c}{Flower-VGG}&\multicolumn{2}{c}{CUB}\\
    \cmidrule(r){3-4}
    \cmidrule(r){5-6}
    \cmidrule(r){7-8}
    \cmidrule(r){9-10}
     & &\textbf{5-way 1-shot} &\textbf{5-way 5-shot} &\textbf{5-way 1-shot} &\textbf{5-way 5-shot}&\textbf{5-way 1-shot} &\textbf{5-way 5-shot}&\textbf{5-way 1-shot} &\textbf{5-way 5-shot} \\
    \midrule
    \multirow{7}{*}{\textbf{FT}}
    & Full Finetuning & 20.02  & 20.32 &  20.07 & 20.01 &  20.00  & 20.08 &  20.01  & 20.03  \\
    & Linear-probe &  76.92 &92.93  & 81.28  &92.95  &  85.12  &96.71  &  78.76  &94.88   \\
    & Lora + Linear & 76.44&  94.85  & 79.20 &  93.60  & 83.17 &  96.57  & 76.39 & 95.43\\
    & $\text{P}>\text{M}>\text{F}$& 77.45  &94.92  &79.31   &93.02  &84.53    &96.46  &77.42    &96.41   \\
    & Loras Avg + Linear & 78.35  & 95.03 & 79.97  & 93.61 &  85.00  & 96.64 &   78.96 &  95.31 \\
    & MOLE&  78.62 &95.23  &79.41   &93.43  &85.12    &96.43  &79.02    &95.38   \\
    & LoraHub &  79.48 &95.36  & 80.12  &  93.93&   85.63 & 96.69 &   79.54 &  95.48 \\
    \midrule
    \multirow{8}{*}{\textbf{FTF}}
        & NN & 75.69  &91.91  & 78.38  &92.55  &  86.47  &96.62  &  77.71  &93.99   \\
    & Loras Avg + NN&  77.05 & 92.56 & 79.63  & 92.60 &  84.21  & 96.35 &  76.32  & 93.61  \\
    & CAML &  78.02 & 93.23 &  80.83 & 93.14 &   85.35 & 96.54 &  78.02  &  94.12 \\
    \cmidrule(r){2-10}
    & \textbf{LoRA Recycle } & 87.37&95.93  & 84.65  &95.03  & \textbf{91.92}$^{\textcolor{blue}{(+6.29\%)}}_{\textcolor{blue}{(+5.45\%)}}$  &97.65  &  \textbf{85.81}$^{\textcolor{blue}{(+6.27\%)}}_{\textcolor{blue}{(+7.79\%)}}$  & \textbf{95.95}$^{\textcolor{blue}{(+0.47\%)}}_{\textcolor{blue}{(+1.83\%)}}$  \\
    & \textbf{LoRA Recycle$_{\textbf{25}}$ }& 87.91  &96.09  &  84.93 & 95.06 &  90.49  & \textbf{97.73}$^{\textcolor{blue}{(+1.02\%)}}_{\textcolor{blue}{(+1.11\%)}}$ & 84.61   & 95.73  \\
    & \textbf{LoRA Recycle$_{\textbf{50}}$ }&  \textbf{88.41}$^{\textcolor{blue}{(+8.93\%)}}_{\textcolor{blue}{(+10.39\%)}}$ & \textbf{96.12}$^{\textcolor{blue}{(+0.76\%)}}_{\textcolor{blue}{(+2.89\%)}}$ &  \textbf{85.61}$^{\textcolor{blue}{(+4.33\%)}}_{\textcolor{blue}{(+4.78\%)}}$ & \textbf{95.33}$^{\textcolor{blue}{(+1.40\%)}}_{\textcolor{blue}{(+2.19\%)}}$ &  90.29  & 97.52 & 84.85   & 95.57  \\
    & \textbf{LoRA Recycle$_{\textbf{75}}$ }&  85.99 & 95.41 &  83.75 & 94.56 & 89.89   & 97.72 & 84.09   & 95.27 \\
     \bottomrule
  \end{tabular}}
  \label{tab:single32}
\end{table*}

\begin{figure*}[htbp]
\vspace{-0.5cm}
  \centering
    \includegraphics[width=0.99\linewidth]{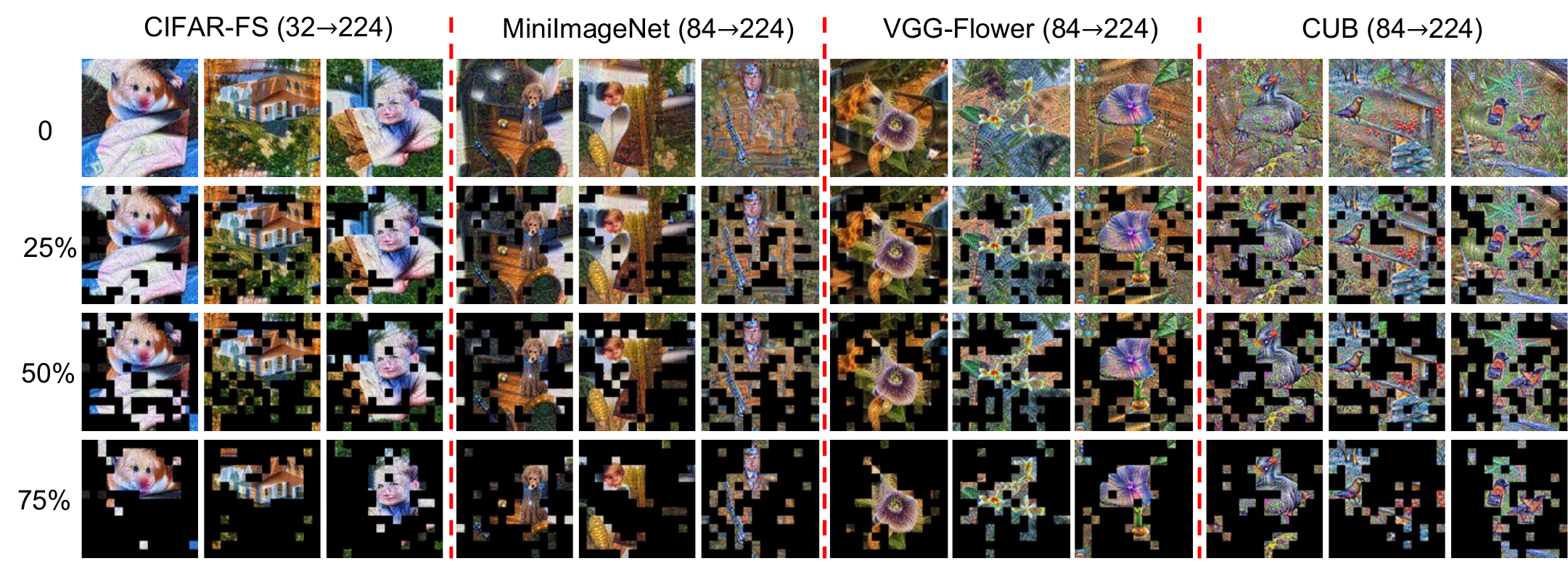}
   \vspace{-0.2cm}
   \caption{Visualization of masked generated images at varying sparsity levels. (32 $\rightarrow$ 224) denotes the original training images' resolution is 32 $\times$ 32 while we can reconstruct images with a higher resolution of 224 $\times$ 224.}
   \label{fig:vis_ratio}
   \vspace{-0.1cm}
\end{figure*}

\noindent
\textbf{Visualization of masked generated images at varying sparsity levels.}\ \cref{fig:vis_ratio} illustrates generated images masked at varying sparsity levels. As we can see, only a subset of tokens carry meaningful semantic information and contribute to the final predictions, while the rest often represent noise, constructed as hallucinations of the VFM's misinterpretations. 
Our method can effectively filter out those noisy tokens and preserve the meaningful tokens, thus effectively preventing  VFM from overfitting to irrelevant noise.

\begin{figure*}[htbb]
  \centering
    \includegraphics[width=0.8\linewidth]{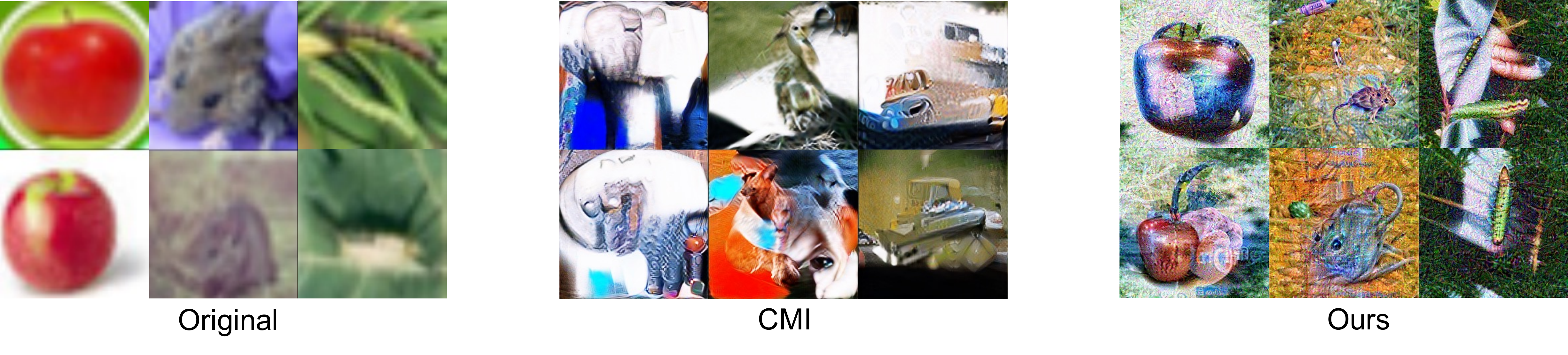}
   \vspace{-0.2cm}
   \caption{Comparison with SOTA model inversion approach. Our model inversion approach surpasses the current SOTA method CMI \citep{fang2021contrastive}, delivering superior image quality with greater efficiency. }
   \label{fig:vis_com}
   \vspace{-0.1cm}
\end{figure*}

\noindent
\textbf{Comparison with SOTA model inversion approach.}\ \cref{fig:vis_com} illustrates that the quality of our model inversion approach  surpasses current state-of-the-art (SOTA) methods like CMI \citep{fang2021contrastive}, which typically produce simpler, lower-resolution images from shallow pre-trained models. Our approach excels in three key areas: (i) quality, producing higher fidelity images; (ii) resolution, capable of generating complex images  with higher resolutions of 224 $\times$ 224; and (iii) efficiency, with our double-efficient mechanism significantly accelerating the model inversion process. Moreover, our work investigates the inversion from transformer-based models, whereas existing methods mainly concentrate on convolutional architectures such as ResNet.

\begin{figure*}[!h]
  \centering
    \includegraphics[width=0.95\linewidth]{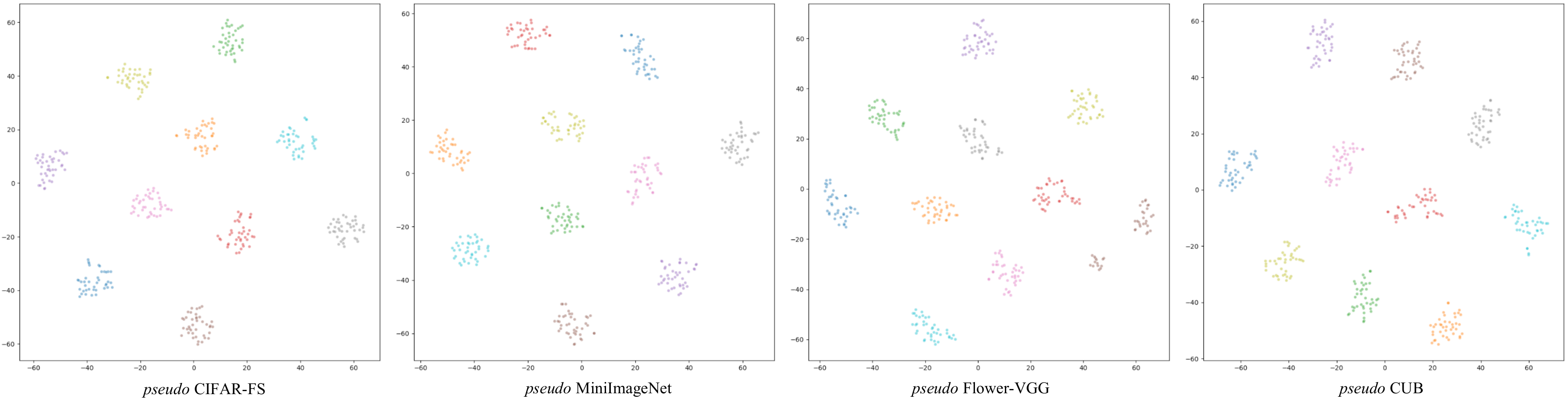}
   \caption{T-SNE visualization of generated images. Our model inversion approach successfully inverts the essential discriminative features, which is beneficial to the following meta-learning.}
   \label{fig:tsne}
   \vspace{-0.2cm}
\end{figure*}

\noindent
\textbf{T-SNE visualization.}\ \cref{fig:tsne} presents the t-SNE visualizations of  images generated from LoRAs pre-tuned on diverse datasets, including CIFAR-FS; MiniImageNet, VGG-Flower, and CUB. Our model inversion approach successfully inverts the essential discriminative features. 

\noindent
\textbf{Effect of the naturalness prior.}\ \cref{fig:bn} shows the efficacy of the regularization term $\mathcal{R}_{\rm BN}$ in \cref{eq:inversion} to enhance the realism of images by enriching natural color and smoothing noise. We set the coefficient $\alpha_{\rm BN}$ as 0.01.

\section{Preliminary of Vision Transformers (ViTs)}
\label{app:preliminary}

\noindent
\textbf{Preliminary of ViTs.} Here, we discuss the operational mechanism behind ViTs. ViTs initially divide the input image ${\boldsymbol{X}^{\rm I}}$ belonging to the space $\mathbb{R}^{H\times W \times C}$ into $n+1$ distinct, non-overlapping patches. These patches are then transformed into $n+1$ tokens, denoted as ${\boldsymbol{X}^{\rm I}}=[\boldsymbol{x}_{\texttt{[CLS]}},\boldsymbol{x}_1,...,\boldsymbol{x}_n]$ where $\boldsymbol{x}_i \in \mathbb{R}^D$. The class token, $\boldsymbol{x}_{\texttt{[CLS]}}$, is prepended to these image tokens to facilitate the classification task. To integrate positional relationships, learnable position encodings are added to all tokens. 
These tokens are then processed through multiple ViT layers, which are composed of multi-head self-attention (MHSA) modules and feed-forward networks (FFN). Within each MHSA, the token set $\boldsymbol{X}^{\rm I}$ undergoes the  transformation into three distinct matrices: the query $\boldsymbol{Q}$, key $\boldsymbol{K}$, and value $\boldsymbol{V}$ matrices. The formulation of the attention mechanism is given by
\begin{equation}
\text{Attention}(\boldsymbol{Q}, \boldsymbol{K}, \boldsymbol{V}) = \text{Softmax}\left(\frac{\boldsymbol{Q} \boldsymbol{K}^T}{\sqrt{d}}\right) \boldsymbol{V},
\label{eq:attention}
\end{equation}
where $d$ represents the dimension of the query vectors within $\boldsymbol{Q}$.
We define $\boldsymbol{A}$ as the square matrix representing the attention weights across all token pairs, calculated as $\boldsymbol{A} = \text{Softmax}\left(\frac{\boldsymbol{Q} \boldsymbol{K}^T}{\sqrt{d}}\right)$, with dimensions $\mathbb{R}^{(n+1)\times(n+1)}$. Specifically, $\boldsymbol{a}_{i}$, which is the $i^{\rm th}$ row of $\boldsymbol{A}$, signifies the attention weights of token $\boldsymbol{x}_i$ with respect to all tokens. Particularly, $\boldsymbol{a}_{\texttt{[CLS]}}$ refers to $\boldsymbol{a}_{0}$. 
Based on \cref{eq:attention}, the $i^{\rm th}$ output token can be viewed as a linear combination of all tokens' value vectors $[\boldsymbol{v}_{\texttt{[CLS]}}, \boldsymbol{v}_{1},...,\boldsymbol{v}_{L}] $, weighted by $\boldsymbol{a}_{i}$.
These output tokens are subsequently forwarded to the FFN, which consists of two linear layers and an  activation function.
At the final ViT layer, the class token $\boldsymbol{x}_{\texttt{[CLS]}}$, summarizing the global image representation, is utilized as the classifier's input to predict the image's classification probability distribution.

\noindent
\textbf{Computational complexity of ViTs.} Given an image split into $N$ patches, each with an embedding dimension of $D$, the computational complexities of self-attention (SA) and feed-forward network (FFN) in ViTs are :
\begin{equation}
    \begin{aligned}
O({\rm S A})=3 N D^2+2 N^2 D, \ \ 
O({\rm F F N})=8 N D^2.
\end{aligned}
\end{equation}
Since the complexities of SA and FFN scale respectively quadratically and linearly with $N$, our proposed double-efficient mechanism (see \cref{sec:double}) significantly reduces the computational complexity by reducing the number of tokens.

\section{Hyperparameter Selection and Sensitivity Analysis}
\label{app:hyperparameter}
In this section, we detail the selection of hyperparameters and conduct a sensitivity analysis on key hyperparameters. Generally speaking, We base our hyperparameter values on reference works and perform grid searches within the relevant ranges to identify the optimal configurations.

For the learning rate in LoRA Inversion, we refer to the settings from prior work \citep{yin2020dreaming}, and perform a grid search over the range \([0.1, 0.25, 0.5]\). Similarly, for the learning rate in the meta-learning stage, we adopt values from the literature \citep{snell2017prototypical} and conduct a grid search over the range \([0.001, 0.01, 0.1]\). These ranges allow us to identify the optimal configurations.

We further conduct sensitivity analysis of  the hyperparameter $\alpha_{\mathcal{R}}$ in \cref{eq:inversion}, as it controls the balance during the inversion process. To analyze this, we conducted experiments on the CIFAR-FS dataset in both 5-way 1-shot and 5-way 5-shot settings. \cref{tab:hyperparameter} shows the results, where we varied the value of $\alpha_{\mathcal{R}}$ to observe its effect on accuracy. Our sensitivity analysis reveals that our framework is not very sensitive to changes in $\alpha_{\mathcal{R}}$, although there are some variations among different $\alpha_{\mathcal{R}}$ values. This stability simplifies the hyperparameter tuning process, making our framework easier to apply in real-world applications.

\begin{table}[h]
  \centering
   \caption{Sensitivity analysis of $\alpha_{\mathcal{R}}$ in \cref{eq:inversion} on CIFAR-FS.} 
   \vspace{-0.3cm}
  \scalebox{0.8}{\begin{tabular}{ccc}
    \toprule
     \textbf{Hyperparameter} & \textbf{5-way 1-shot} & \textbf{5-way 5-shot} \\
     \midrule
     0.1&89.35&96.39\\
     0.01&89.70&96.69\\
     0.001&88.83&95.76\\
     \bottomrule
  \end{tabular}}
  \label{tab:hyperparameter}
\end{table}

\section{Implementation Details of baselines}
\label{app:baselines}
Here, we provide detailed implementation details for the baselines used in our paper.. 
\begin{itemize}[leftmargin=20pt]
    \item \textbf{Fine-tuning baselines.} ``Full Fine-Tuning" updates the entire model on the target task via gradient descent. ``Linear probe" only updates the classification head. ``LoRA + Linear \citep{hu2021LoRA}" updates the layer-wise rank decomposition matrices and the classification head. For fine-tuning, we select the best results from learning rates $[0.1, 0.01, 0.001]$. For LoRA, we set the rank to 4.
    \item \textbf{Multi-LoRAs composition baselines.} ``LoRAs Avg'' refers to averaging all given pre-tuned LoRAs into a single LoRA, which can be further fine-tuned with the classification head (``LoRAs Avg +  Linear") or directly make inference via Nearest Neighbour (``LoRAs Avg +  NN") without fine-tuning. ``LoRAHub \citep{huang2023LoRAhub}" takes a further step which obtains a single LoRA by a weighted sum of given pre-tuned LoRAs, where the weight values are fine-tuned on the target task. ``MOLE \citep{chen2024llava}" fine-tunes a
learnable gating function to composing the outputs of different LoRAs. For LoRAHub, we use a gradient-free approach to fine-tune the coefficients of pre-tuned LoRAs, following the setup in the original paper. For MOLE, we use gradient descent to fine-tune the learnable gating function. We select the best fine-tuning results from learning rates $[0.1, 0.01, 0.001]$.
    \item \textbf{Few-shot adaptation.} The current state-of-the-art baseline, $\text{P} > \text{M} > \text{F}$ \citep{hu2022pushing}, performs few-shot adaptation by stacking three stages: pre-training, meta-training, and fine-tuning. We follow the original paper’s setup and apply data augmentation to the support set of the target tasks. We select the best fine-tuning results from learning rates $[0.1, 0.01, 0.001]$.
    \item \textbf{Fine-tuning-free baselines.} ``Nearest Neighbour (NN)" makes predictions based on the label of the closest class center. "CAML \citep{fifty2023context}" trains a sequence model to simulate the in-context learning of LLMs. Since we do not have real data to train the sequence model, we use generated data generated from pre-tuned LoRAs to train the sequence model. All other settings are consistent with the original paper.
\end{itemize}

\section{More Discussions}
\label{app:discussion}

\noindent
\textbf{Discussions on the inconsistent performance gains across various datasets.}  When we use LoRAs from the dataset the same as the testing dataset (in-domain setting), those LoRAs can provide domain-specific priors. This is particularly useful when the foundation model's pre-training dataset varies from the testing dataset.
The main paper's \cref{tab:single16}  confirms this, showing a higher performance gain on CIFAR-FS (+10.01\%) than other datasets (average +4.98\%). 
The larger disparity between CIFAR-FS and the pre-training dataset is supported by the baseline NN in the main paper's \cref{tab:single16}, showing that directly transferring the foundation model to CIFAR-FS results in a lower accuracy (78.06\%) compared to other testing datasets (average 85.31\%). When we use LoRAs from datasets different from the testing dataset (cross-domain setting), performance gains across datasets are relatively stable, since these LoRAs offer limited useful domain-specific priors for all testing datasets.

\noindent
\textbf{Paradigms for Adaptable Foundation Models}
Several paradigms have been proposed to make large foundation models more adaptable. These paradigms involve combinations among Pre-training (P), Meta-learning (M), Fine-tuning (F)  or PEFT, and In-context learning (I). Here, we provide a discussion over three paradigms, including P$\textgreater$F or P$\textgreater$PEFT, P$\textgreater$M$\textgreater$F and P$\textgreater$M$\textgreater$I. $\textgreater$ indicates the sequence.
Traditional P$\textgreater$F and P$\textgreater$PEFT \citep{fu2023effectiveness,lee2022surgical,sun2019meta} often fail to adapt foundation models to data-limited and real-time applications due to their need for sufficient data and explicit fine-tuning.

An emerging strategy, P$\textgreater$M$\textgreater$F, introduces a meta-learning phase before fine-tuning, preparing the pre-trained model for subsequent fine-tuning. This paradigm has shown promising results in vision \citep{hu2022pushing,cai2020cross}, language \citep{gheini-etal-2023-know,bansal2022meta,hou2022meta} and vision-language \citep{yeh2023meta,najdenkoska2023meta,huang2023diversity} domains.

More recently, the P$\textgreater$M$\textgreater$I paradigm has been proposed in language domains, aiming to acquire more advanced in-context learning ability of LLMs.
For example, LLMs are equipped with the instruction-following ability by meta-training on a broad range of tasks accompanied by instructions \citep{iyer2022opt,chung2022scaling}. MetaICL \citep{min-etal-2022-metaicl} and ICT \citep{chen-etal-2022-meta} explicitly meta-train LLMs to learn to learn in context. 
However, paradigms for fine-tuning-free adaptation in VFMs are less explored, hindered by their inherent in-context learning limitations compared to LLMs.

\noindent
\textbf{Difference between domain generalization and our setting.} Our setting is fundamentally different from domain generalization \cite{wang2022generalizing} in several key aspects:
Domain generalization aims to learn across multiple domains to generalize to an unseen domain. It requires that both known and unseen domains share the same label space. For example, training domain 1 may include real images of cats and dogs, and training domain 2 may include animated images of cats and dogs. Then, the test domain would include paintings of cats and dogs.
Our setting is more challenging, as both the labels and domains for training and testing tasks differ. 

Moreover, the labels in the test tasks are unseen during training. For example, in our setting, Task 1 might involve real images of cats and dogs, Task 2 might involve animated images of tigers and lions, and the test task could involve paintings of chairs and tables.

Additionally, unlike domain generalization, our setting emphasizes a few-shot scenario in test tasks and does not require original data in training tasks.

\noindent
\textbf{Discussions on data-free knowledge distillation (DFKD).} Data-Free Knowledge Distillation (DFKD) \cite{yu2023data,tran2024nayer,wang2024confounded,liu2024small,shao2023data,fang2022up,tran2024large,husparse} facilitates the transfer of knowledge from a large pre-trained teacher model to a smaller, more efficient student model without requiring access to the original training data. This methodology is particularly significant in scenarios where privacy or ethical concerns limit data accessibility. DFKD approaches such as DeepInversion \cite{yin2020dreaming} and CMI \cite{fang2021contrastive} synthesize images by utilizing teacher model statistics and classification objectives, and the synthesized images are used to perform knowledge transfer. Recently, ABD \cite{hong2023revisiting} investigates potential security vulnerabilities in DFKD, with a focus on backdoor threats. DFKD techniques have also been applied to areas such as federated learning \cite{luo2023dfrd} and model quantization \cite{li2022psaqvit,li2023psaq}.

Unlike DFKD, which primarily employs inverted data to distill knowledge from a single teacher model, our study introduces a meta-learning framework that harnesses inverted data across multiple teacher models. Moreover, instead of transferring task-specific knowledge, our framework aims to learn generalizable prior \textit{meta-knowledge} \cite{finn2017model}, which can be rapidly adapted to new tasks. Lastly, we propose a novel dual-efficiency mechanism that accelerates both the data inversion and meta-training processes. This contribution not only speed up our meta-learning framework but also holds potential for improving the inversion efficiency in standard DFKD methods.

\section{Rethinking Existing Data-Free Meta-Learning Methods}
\label{app:rethinking}
\textbf{Limited scalability to large-scale models.} Current data-free meta-learning (DFML) methodologies, as discussed in \citep{wang2022metalearning, hu2023architecture, hu2023learning}, predominantly focus on leveraging small-scale pre-trained models and meta-learners, such as four-layer CNNs or ResNet12. A critical limitation of these approaches is their inability to scale up to larger models, particularly those based on transformer architectures. This scalability issue substantially hinders their practical application in complex, real-world scenarios. For instance, \citep{wang2022metalearning} employs a hyper-network with all pre-trained models as inputs and outputs a single fused model. The efficiency of this method declines significantly when outputting all parameters of larger models, given the hyper-network's extensive input and output dimensions. Similarly, inversion-based DFML methods, such as those in \citep{hu2023architecture, hu2023learning}, rely on meta-training a meta-learner with  data inverted from pre-trained models. The model inversion process becomes inefficient for large-scale models. The following meta-training process often necessitates the computation of Hessian matrices for second-order derivatives \citep{nichol2018first}, which becomes exceedingly resource-intensive for large-scale models.

\noindent
\textbf{Inefficiency issues.} Beyond scalability challenges, inversion-based methods like \citep{hu2023architecture, hu2023learning} are plagued by the inefficiency of the model inversion processes. These methods typically involve iterative forward and backward optimizations, leading to significant computational and storage costs when applied to large-scale models.

\noindent
\textbf{Addressing these issues: our contributions.} Our Double-Efficient Data-Free Meta-Learning framework presents a novel and efficient solution that is scalable for transformer-based foundation models. To overcome existing limitations,
(i) we propose a data-free
meta-learning framework, which is specifically designed for large-scale VFM. We only meta-train meta-LoRA, constituting only  0.14M
parameters (merely 0.1\% relative to the VFM).
(ii) We propose a meta-learning objective, as outlined in \cref{eq:LoRAhubDistillation}, that avoids the resource-intensive computation of Hessian matrices. This is achieved as our fine-tuning-free adaptation in the inner loop does not require gradient computations.
(iii) We propose a double-efficient mechanism that significantly speeds up the meta-training processes while maintaining comparable or enhanced performance. 
Our approach not only addresses the limited scalability and inefficiency issues of existing DFML methods, but also inspires more interactions between meta-learning and foundation models.

\end{appendix}